\documentclass[conference,10pt]{IEEEtran}

\usepackage[letterpaper, top=60pt, bottom=43pt, left=48pt, right=48pt]{geometry}
\usepackage{tcolorbox}
\usepackage[ruled,linesnumbered]{algorithm2e}
\usepackage{bm}
\usepackage{stackrel}
\usepackage{subfigure}
\usepackage{epsf}
\usepackage{amsmath,amssymb}
\usepackage{graphicx}
\usepackage{hyperref}
\usepackage{color}
\usepackage{cite}
\usepackage{multirow,tabularx}
\usepackage{ifthen}
\usepackage{epstopdf}
\usepackage{mathtools}
\usepackage{xcolor} % for colored boxes
\input{epstopdf.sty}
\usepackage{dblfloatfix}

\usepackage{times}

\input{epsf.sty}

\usepackage{booktabs} % for professional tables
\usepackage{capt-of}
\usepackage{microtype}
\usepackage{hyperref}
\usepackage{xr}

\usepackage{lipsum}
\usepackage{optidef}

\newcommand{\ralOmit}[1]{#1}
\newcommand{\ralAdd}[1]{}

\makeatletter
\newcommand{\printfnsymbol}[1]{%
	\textsuperscript{\@fnsymbol{#1}}%
}
\makeatother

\newcommand{\bdmath}{\begin{dmath}}
\newcommand{\edmath}{\end{dmath}}
\newcommand{\beq}{\begin{equation}}
\newcommand{\eeq}{\end{equation}}
\newcommand{\bdm}{\begin{displaymath}}
\newcommand{\edm}{\end{displaymath}}
\newcommand{\bea}{\begin{eqnarray}}
\newcommand{\eea}{\end{eqnarray}}
\newcommand{\beal}{\beq \begin{array}{ll}}
\newcommand{\eeal}{\end{array} \eeq}
\newcommand{\beas}{\begin{eqnarray*}}
\newcommand{\eeas}{\end{eqnarray*}}
\newcommand{\ba}{\begin{array}}
\newcommand{\ea}{\end{array}}
\newcommand{\bit}{\begin{itemize}}
\newcommand{\eit}{\end{itemize}}
\newcommand{\ben}{\begin{enumerate}}
\newcommand{\een}{\end{enumerate}}

\newcommand{\calG}{{\cal G}}

\newcommand{\etal}{\emph{et~al.}\xspace}

\newcommand{\eg}{\emph{e.g.,}\xspace}
\newcommand{\ie}{\emph{i.e.,}\xspace}

\newcommand{\M}[1]{{\bm #1}} % Face for matrices

\newcommand{\hide}[1]{}

\newcommand{\hiddenText}{{\color{gray} hidden text.}}
\newcommand{\hideWithText}[1]{\hiddenText}

\newcommand{\MG}{\M{G}}

\newcommand{\MI}{\M{I}}

\newcommand{\blue}[1]{{\color{blue}#1}}
\newcommand{\green}[1]{{\color{green}#1}}
\newcommand{\red}[1]{{\color{red}#1}}

\newcommand{\linkToPdf}[1]{\href{#1}{\blue{(pdf)}}}
\newcommand{\linkToPpt}[1]{\href{#1}{\blue{(ppt)}}}
\newcommand{\linkToCode}[1]{\href{#1}{\blue{(code)}}}
\newcommand{\linkToWeb}[1]{\href{#1}{\blue{(web)}}}
\newcommand{\linkToVideo}[1]{\href{#1}{\blue{(video)}}}
\newcommand{\linkToMedia}[1]{\href{#1}{\blue{(media)}}}
\newcommand{\award}[1]{\xspace} % {{\red{#1}}} % omit awards

\graphicspath{{./figures/}}

\begin{document}

%\title{Do Object Shape and Pose Estimation Models Enable Better Grasps?}
%\title{Object Perception for Grasping: Does it Work?}
\title{Object Pose and Shape Estimation \\ for Grasping: Does it Work?}

%\author{Pavan Karke\printfnsymbol{1}\thanks{\printfnsymbol{1} equal contribution}, Kushal Shah\printfnsymbol{1}, Gaurav Singh, Md Faizal Karim, K Madhava Krishna, and Rajat Talak}
\ralOmit{
\author{Pavan Karke$^{1\ast}$, Kushal Shah$^{1\ast}$, Gaurav Singh$^{1}$, Md Faizal Karim$^{1}$, K Madhava Krishna$^{1}$, and Rajat Talak$^{2}$ \vspace{1mm}\\

$^\ast$Equal Contributions, $^{1}$Robotics Research Center, IIIT Hyderabad, $^{2}$National University of Singapore}
}
\ralAdd{\author{Anonymous}}

\IEEEaftertitletext{\vspace{-0.6\baselineskip}}

% \maketitle

%% Macros used in the paper

\newcommand{\anyGrasp}{\texttt{AnyGrasp}\xspace}
\newcommand{\crisp}{\texttt{CRISP}\xspace}
\newcommand{\crispLong}{category-agnostic pose and shape estimation\xspace}
\newcommand{\sceneCompleteShort}{\texttt{SceneComplete}\xspace}
\newcommand{\sceneComplete}{\text{Scene Complete}\xspace}
\newcommand{\sC}{\texttt{SC}\xspace}
\newcommand{\samThreeD}{\texttt{SAM3D}\xspace}

\newcommand{\antipodal}{\texttt{Antipodal}\xspace}

\newcommand{\crispGrasp}{\texttt{CRISP-Grasp}\xspace}
\newcommand{\scGrasp}{\texttt{SC-Grasp}\xspace}
\newcommand{\samThreeDGrasp}{\texttt{SAM3D-Grasp}\xspace}
\newcommand{\lerfToGo}{\texttt{LERF-TOGO}\xspace}

\newcommand{\anyGraspFilterCRISP}{\text{\anyGrasp\!\!+\crisp}\xspace}
\newcommand{\anyGraspFilterSC}{\text{\anyGrasp\!\!+\sC}\xspace}

\newcommand{\graspSpace}{\ensuremath{\calG}\xspace}

\definecolor{mygreen}{HTML}{599E94}
\definecolor{mybrown}{HTML}{6D4B4B}
\definecolor{myred}{HTML}{E27C7C}

%\maketitle
%
\twocolumn[{
\renewcommand\twocolumn[1][]{#1}
\maketitle
%
%\begin{figure*}[h]
\begin{center}
% \vspace{5mm}
 
 \centering
 \ralAdd{\includegraphics[width=0.9\textwidth]{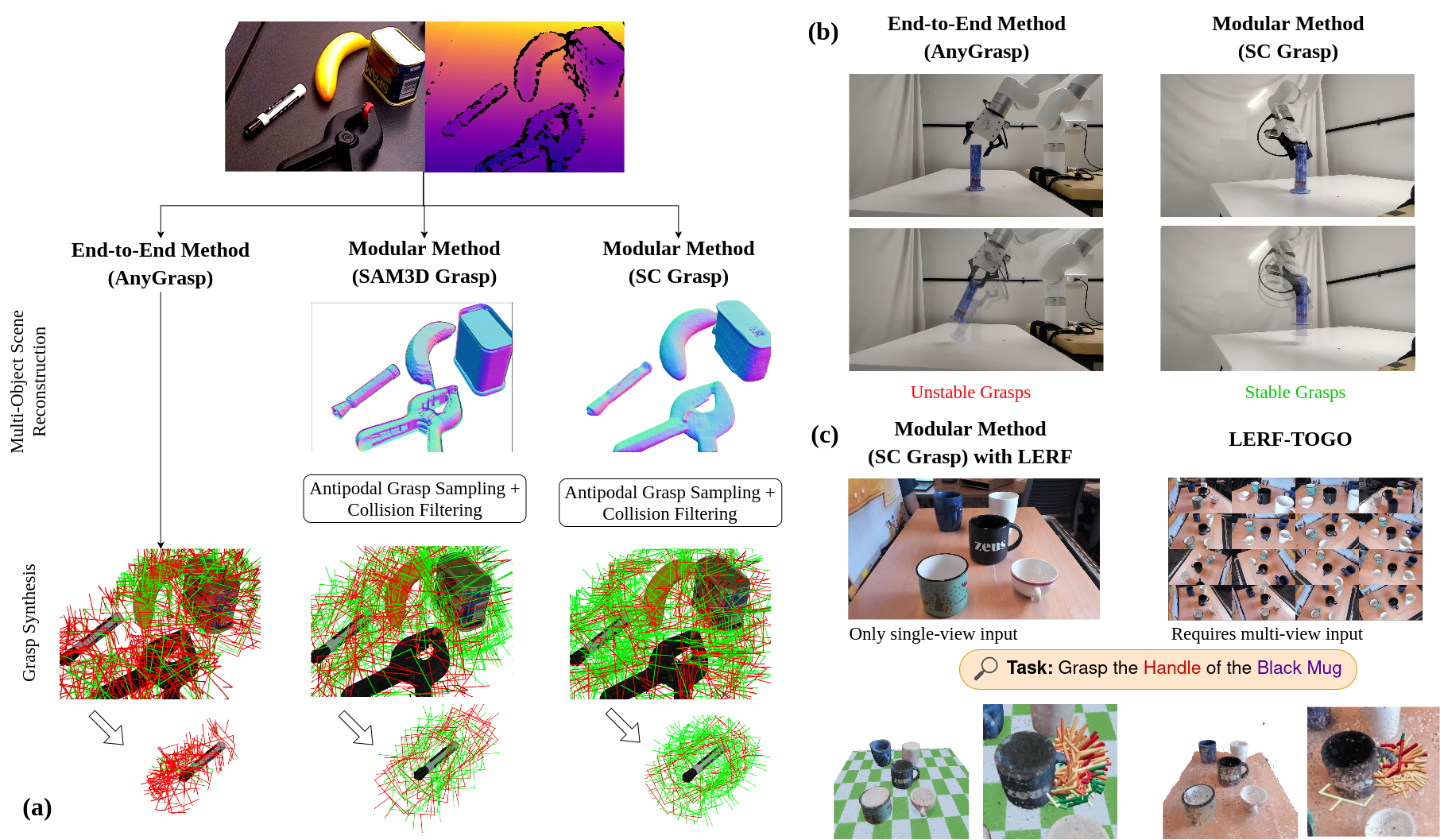}}
 \ralOmit{\includegraphics[width=0.99\textwidth]{figures/FinalOpenSAM.png}}
 \label{fig:teaser}
 \ralAdd{\vspace{-4mm}}
 \ralOmit{\vspace{-3mm}}
\end{center}

\captionof{figure}{
%Fig. Opening: 
(a) We implement and analyze four methods for grasp synthesis (only three shown): a state-of-the-art, End-to-End Method (\anyGrasp~\cite{Fang23tro-anyGrasp}), and three Modular Methods (\samThreeDGrasp, \crispGrasp and \scGrasp), which first estimate the object pose and shape for all objects in the scene and then synthesize grasps using antipodal sampling. 
%We use recent state-of-the-art models CRISP~\cite{Shi25cvpr-CRISP} and SceneComplete (SC)~\cite{Agarwal24arxiv-sceneComplete} for pose and shape estimation.
We observe that Modular Methods synthesize more successful grasps (\green{green}) than the End-to-End Method (unsuccessful grasps: \red{red}). %; success here is defined using the force closure and collision metric.  
%The modular methods also synthesize more number of grasps even for small objects.
(b) The End-to-End Method synthesize more unstable grasps compared to Modular Methods in our real-world experiments.
(c) The Modular Methods are augmented with vision-language models to yield language-conditioned grasps from just single-view RGB-D input. We observe comparable performance with the state-of-the-art baseline \lerfToGo~\cite{Rashid23corl-lerfToGo} that uses multiple views. 
}
%
%Figure: This is an opening figure. \Madhav{It is not straightaway clear for me what is being conveyed here. How are we showing the proposed Two Stage is indeed doing better? Pavan, Kushal to see how to get this idea across }
%\vspace{5mm}
\vspace{3mm}
%\end{figure*}
}]

%!TEX root = ../main.tex

\begin{abstract}
%
%
%\red{Story 1}
%Object perception, the problem of estimating complete geometry of the object from a partial-view, has seen key advancements lately. 
The problem of object pose and shape estimation has seen key advancements lately. 
%
%The problem of object pose and shape estimation has seen 
%The problem of estimating pose and shape of an object from a single RGB(-D) image has seen progress lately.
Encoder-decoder (\eg SAM3D, LRM, CRISP) and diffusion-based models (\eg InstantMesh, Zero123, SceneComplete) have shown category-agnostic shape encoding capacity and open-set generalizability. 
%Diffusion models, utilize novel-view synthesis, and show open-set generalizability, whereas direct regression models show category-agnostic shape encoding.   
%These methods are able to reconstruct the 3D object geometry from a single image without any prior CAD model. 
%
%
In this work, we ask the question: 

\vspace{1mm}
\emph{Are the object pose and shape estimation methods mature enough, such that when used with antipodal grasp sampling, can outperform the end-to-end grasp synthesis methods?}
\vspace{2mm}

%
%The model-free object pose and shape estimation methods, augmented with classical grasp sampling, 
%The classical grasp sampling methods sample grasps on an object mesh using geometric and physical constraints. 
%
\noindent We explore this question in detail by scoping our study to parallel-jaw grippers, 7-DoF grasps, and single-view RGB(-D) image as input. 
We implement and compare a state-of-the-art, end-to-end grasp synthesis method and three modular methods, which first estimate the object pose and shape for all objects in the scene, and generates grasps using antipodal sampling.
We observe that the modular methods outperform the end-to-end method in all our experiments. 
The modular methods are able to synthesize plenty of grasps, even for small objects, where the end-to-end methods fail.
%
% A trivial augmentation of object pose and shape estimation into the end-to-end-grasp synthesis yields only marginal performance gains; thus, a non-trivial solution is called for when aiming for the best of both worlds.  %
%
The effectiveness of the modular methods is contingent on the accuracy of the pose and shape estimation, and suffers partial degradation in cluttered scenes --- a limitation of the existing pose and shape estimation methods.
%
%We see that a trivial augmentation of object pose and shape estimation into the end-to-end grasp synthesis model yields only marginal performance gains in real-world setting; thus, a non-trivial solution is called for when aiming for the best of both worlds. 
%
We also analyze the failure modes and run-times for the three modular methods, which use two different ways of object pose and shape estimation: one based on an encoder-decoder model, while another a diffusion model.
%%
%
%
%
%%
%We see that the modular methods are able to synthesize plenty of grasps, whereas the number synthesized by the end-to-end method is dependent on the object size and degrade its performance. 
%%
%On the other hand, we see that the modular methods suffer in cluttered environments, indicating that the pose and shape estimation models are not yet robust to occlusions.
%%
%We see that a trivial augmentation of object perception into the end-to-end grasp synthesis model yields only marginal performance gains in real-world setting; thus, a non-trivial solution is called for when aiming for the best of both worlds. 
%%
%We also investigate and find different grasp failure modes for the two modular methods that we implement -- one uses the encoder-decoder model for pose and shape estimation, whereas the second uses a diffusion model. 
%%
%A failing grasp on the encoder-decoder-based modular method is unlikely due to a shape estimation error ($6.4\%$), whereas it is more likely ($29.6\%$) for the diffusion model-based modular method.
%
%We examine the run-times for the three grasp synthesis methods and bring out important lessons.
%
%
Finally, we demonstrate that the single-view object pose and shape estimation methods can be augmented with vision-language models to yield language-conditioned grasps from just single-view RGB-D image as input. We notice comparable performance to the state-of-the-art \lerfToGo baseline.

\end{abstract} 
\vspace{2mm}


%!TEX root = ../main.tex

\section{Introduction}
\label{sec:intro}

The problem of estimating the complete geometry of an object from a single- or partial-view is very important in robotics. 
It can help robots navigate a scene as well as grasp and manipulate objects with very few observations. 
It is a capability that mobile manipulators aught to have. % but do not. \Madhav{This statement can be rephrased or dropped. "Imbuing manipulators with such abilities greatly enhances their performance" }\RT{I know it is redundant; but I think it is fine. Would be good to show relevance to mobile manipulation community.} 
Classical works on grasping, for instance, relied on the assumption that such a complete object geometry is available for grasp synthesis~\cite{Shimoga96ijrr-robustGraspSurvey,Bicchi00icra-graspingReview}. Knowing the complete geometry, material, and physical properties, the grasp synthesis problem was framed as an optimization problem to seek the grasp that is most likely to succeed~\cite{Bicchi00icra-graspingReview,Bohg14tro-graspSurvey}. Recent works have used these physics constraints to develop grasp sampling approaches given a 3D mesh of the object~\cite{Mahler17rss-dexnet2,Zhai22ral-da2dataset,Eppner19isrr-billionWaystoGrasp}. 
%
%and optimization approaches that sought the grasps that were most likely to succeed.  
% However, estimating the complete geometry, from single- or partial-view, is an underspecified problem and cannot be solved without sufficient induced priors. 
%It was one of the first problems to be 
%

%% Papers
% Category-level
% RePoNet~\cite{Fu22nips-wild6d}, CASS~\cite{Chen20iccv-learningCanonicalShape}, SGPA~\cite{Chen21iccv-sgpa}, SPD~\cite{Tian20eccv-SPD}, DualPoseNet~\cite{Lin21iccv-dualposenet}, FSD~\cite{Lunayach24icra-FSD}. 
%
%
% Encoder-Decoder Models
%\eg CRISP~\cite{Shi25cvpr-CRISP}, LRM~\cite{Hong24iclr-lrm}, ZeroShape~\cite{Huang24cvpr-zeroShape}, MCC~\cite{Wu23cvpr-mvCompressiveCoding}, and SS3D~\cite{Alwala22cvpr-SS3D}
%
% Diffusion Models
%InstantMesh~\cite{Xu24arxiv-instantmesh}, Zero123~\cite{Liu23iccv-zero1to3}, ShapE~\cite{Jun23arxiv-shape}, NeuralLift 360~\cite{Xu23cvpr-neuralLift360}, RealFusion~\cite{MelasKyriazi23cvpr-realFusion}

%This problem, however, has seen key advancements lately. 
The problem of estimating the complete geometry of objects (\ie shape and pose of the object) given a single-view RGB(-D) image has seen key advancements 
\ralAdd{lately.}
\ralOmit{lately~\cite{Wang21cvpr-GDRNetGeometryGuided, Wang19cvpr-nocs, Liu25pami-GDRNPP, Fu22nips-wild6d, Chen21iccv-sgpa, Lin21iccv-dualposenet, Wu23cvpr-mvCompressiveCoding,Hong24iclr-lrm,Shi25cvpr-CRISP,Alwala22cvpr-SS3D,Huang24cvpr-zeroShape,Xu24arxiv-instantmesh,Liu23iccv-zero1to3,Jun23arxiv-shape,Xu23cvpr-neuralLift360,MelasKyriazi23cvpr-realFusion,Agarwal24arxiv-sceneComplete}.} 
Encoder-decoder models have shown the capability to estimate the pose and shape of an object, across object categories~\cite{Shi25cvpr-CRISP,Wu23cvpr-mvCompressiveCoding,Hong24iclr-lrm,Alwala22cvpr-SS3D,Huang24cvpr-zeroShape}, and most recently, open-set generalizability~\cite{sam3d-objects} (\ie the capability to estimate the pose and shape of previously unseen objects). 
\ralOmit{Previously, researchers were working on category-level object pose and shape estimation, where the template object shape would be known~\cite{Wang19cvpr-nocs,Wang20eccv-Self6DSelfSupervised,Zhang24arxiv-shapeICP,Fu22nips-wild6d,Chen20iccv-learningCanonicalShape,Chen21iccv-sgpa,Tian20eccv-SPD,Lin21iccv-dualposenet,Lunayach24icra-FSD}. It was believed that such a category-agnostic generalization would not be possible.} 
Diffusion models have also shown open-set generalizability~\cite{Xu24arxiv-instantmesh,Liu23iccv-zero1to3,Jun23arxiv-shape,Xu23cvpr-neuralLift360,MelasKyriazi23cvpr-realFusion,Agarwal24arxiv-sceneComplete}.
These models leverage the progress on novel-view synthesis~\cite{Watson22arxiv-novelView}, which once framed as a pose conditioned image-to-image generation problem and trained using a large diffusion model shows zero-shot generalizability.

%
%As a consequence, 
Today's grasp synthesis approaches are dominated by end-to-end regression methods which regress the grasp pose directly from a single- or partial-view of a scene~\cite{Mahler17rss-dexnet2,Fang20cvpr-graspnet1B,Song20ral-graspingWild,Sundermeyer21icra-contactGraspNet,Gou21icra-rgbMatters,Fang23tro-anyGrasp,Newbury23tro-graspReview}. 
These methods were initially inspired by the success of the end-to-end deep learning methods in computer vision~\cite{Bohg14tro-graspSurvey,Newbury23tro-graspReview}. They have now shown remarkable capability in terms of increased accuracy, generalizability to unseen objects, and robustness compared to traditional machine learning or analytical methods. End-to-end methods have also enabled faster run times for real-time operations, unlike sampling-based methods~\cite{Newbury23tro-graspReview}.
%
% The end-to-end methods are now being extended to task-oriented grasping~\cite{Fang20ijrr-tog,Rashid23corl-lerfToGo}, language-driven grasping~\cite{Vuong24cvpr-languageGrasp,Rashid23corl-lerfToGo}, dynamic object grasping~\cite{Fang23tro-anyGrasp}, and cross embodiment grasping (\ie grasp synthesis that works for different robot grippers)~\cite{Wei24icra-droGrasp}.
%seeking to replicate its success for the task of grasp synthesis. 
%
%
%\red{Add history about end-to-end grasping.}
%

\textbf{Contributions.} In this work, we inquire if this recent progress in object pose and shape estimation can be leveraged for grasping. 
%
% In particular, we 
% %In this work, we
% %ask: 
% ask: 
%
% \vspace{3mm}
% \emph{Are the object pose and shape estimation methods mature enough, such that when used with antipodal grasp sampling, can outperform the end-to-end grasp synthesis methods?}
% \vspace{3mm}
% 
% \noindent We explore this question via experimental analysis. 
We explore this question via experimental analysis.
%Limiting \Madhav{Scoping or Bounding our study instead of limiting?}\RT{Done!}
Scoping our study to parallel grippers, 7-DoF grasps, and single-view RGB-D image as input we analyze and compare a state-of-the-art, end-to-end grasp synthesis method and three modular methods, which first estimate object pose and shape of all the objects in the scene, and generate grasps using antipodal sampling~\cite{Mahler17rss-dexnet2,Zhai22ral-da2dataset}. 
We evaluate the grasp synthesis methods with collisions, force closure, and grasp stability. % in real-world experiments. 
We conduct our experiments in a physics simulator, real-world datasets, and real-world grasping experiments with Xarm7 manipulator arm affixed with a parallel-jaw gripper and an Intel RealSense D455 camera. 
We demonstrate that the single-view object pose and shape estimation methods can also be leveraged for language-conditioned grasp synthesis with (qualitatively) comparable performance vis-a-vis a state-of-the-art multi-view baseline.

\section{A Discussion with Related Works}
\label{sec:rel}

\subsection{Object Pose and Shape Estimation}
\label{sec:rel-pose-and-shape}

The earliest works on object pose and shape estimation were motivated by its application to grasping and manipulation. Early works focused on fitting templates (\eg cuboids, ellipsoids, quadrics). Instance-based object pose estimation gained traction for industrial applications where the object shape is known and could be easily retrieved~\cite{Hodan20eccvw-BOPChallenge}. Deep learning approaches showed pivotal role in increasing the accuracy of instance-based pose estimation methods\ralAdd{~\cite{Wang21cvpr-GDRNetGeometryGuided,Wen24cvpr-foundationPose}.}\ralOmit{~\cite{Xiang18rss-posecnn,Wang21cvpr-GDRNetGeometryGuided,Caraffa24eccv-freeze,Moon25cvpr-coop,Wen24cvpr-foundationPose,Nguyen24cvpr-gigaPose,Liu25pami-GDRNPP,Moon24cvpr-genFlow,Labbe22corl-megapose}.} 
The success of these methods prompted researchers to investigate category-level pose and shape estimation task, where an object template shape is assumed to be known\ralAdd{~\cite{Wang19cvpr-nocs,Wang20eccv-Self6DSelfSupervised}.}\ralOmit{~\cite{Wang19cvpr-nocs,Wang20eccv-Self6DSelfSupervised,Zhang24arxiv-shapeICP,Fu22nips-wild6d,Chen20iccv-learningCanonicalShape,Chen21iccv-sgpa,Tian20eccv-SPD,Lin21iccv-dualposenet,Lunayach24icra-FSD}.} These methods estimate the object pose and shape by modifying the template shape to fit the object instance in the input image. 

More recently, there has been a realization that the encoder-decoder models are capable of encoding many object shapes, irrespective of their semantic category~\cite{Shi25cvpr-CRISP,Wu23cvpr-mvCompressiveCoding,Hong24iclr-lrm,Alwala22cvpr-SS3D,Huang24cvpr-zeroShape,sam3d-objects}. 
This has enabled researchers to propose a single deep learning model that can estimate both the shape and pose of the object in the input image. Supervised training on large-scale synthesis data provides the required inductive priors. 
Hong \etal~\cite{Hong24iclr-lrm} and Huang \etal~\cite{Huang24cvpr-zeroShape} proposes a large reconstruction model trained on a large synthetic dataset to enable zero-shot generalization to unseen objects and unseen environments.  
Wu \etal~\cite{Wu23cvpr-mvCompressiveCoding} proposes a way of training an object reconstruction model using monocular videos. 
Shi \etal~\cite{Shi25cvpr-CRISP}, on the other hand, proposes an encoder-decoder model trained on a dataset of objects so that it can show generalization across unseen environments, but known objects. The work shows some zero-shot generalization capability. 
Very recently, SAM3D~\cite{sam3d-objects} released by Meta Research has been able to show zero-shot generalization to unseen objects and unknown environments. 

Diffusion models are a major contenders for zero-shot generalization when it comes to object pose and shape estimation~\cite{Xu24arxiv-instantmesh,Liu23iccv-zero1to3,Jun23arxiv-shape,Xu23cvpr-neuralLift360,MelasKyriazi23cvpr-realFusion,Agarwal24arxiv-sceneComplete}. These models utilize the recent breakthrough in novel-view synthesis, where in you can render a novel view of an object, given a single-view input image and the relative pose transformation. The diffusion models render several views of the single object and train a NeRF or a distance field to capture the object geometry and texture. 
These models have shown remarkable zero-shot generalization properties to unknown objects and unseen environments. More recently, they have been augmented to mitigate against occlusions and used for zero-shot scene reconstruction~\cite{Agarwal24arxiv-sceneComplete}.
A major known drawback of these methods is its high run-time. It takes a few seconds to estimate the pose and shape of an object.

\subsection{Grasp Synthesis}
\label{sec:rel-grasp-synthesis}

Grasp synthesis has been a fundamental problem in robotics and has been investigated for over four decades. 
Early works formulated the problem as a non-linear optimization problem, aiming to minimize a dexterity measure, while the grasp satisfies equilibrium constraints like force closure~\cite{Shimoga96ijrr-robustGraspSurvey,Bicchi00icra-graspingReview}.
Investigation into data-driven approaches began with the availability of Graspit!~\cite{Miller04ram-graspIt} in the mid-2000s, and gained pace with the advancements in deep learning, photorealistic simulators, and physics simulators that could generate synthetic training data~\cite{Bohg14tro-graspSurvey,Newbury23tro-graspReview,Eppner19isrr-billionWaystoGrasp,Mahler17rss-dexnet2,Fang20cvpr-graspnet1B}. 

Many seminal works showed that end-to-end models that directly regress grasp poses could achieve very high success rate, while generalizing to unseen objects and environments~\cite{Mahler17rss-dexnet2,Fang20cvpr-graspnet1B,Song20ral-graspingWild,Sundermeyer21icra-contactGraspNet,Gou21icra-rgbMatters,Fang23tro-anyGrasp}. 
These methods differed mainly by the way they represent the grasp pose and the object.
Sundermeyer \etal~\cite{Sundermeyer21icra-contactGraspNet} showed that 6 DoF grasp can be represented with four degrees of freedom if one uses a depth point on the object as a contact point.
Fang \etal~\cite{Fang23tro-anyGrasp} incorporates center of gravity estimates into its grasp synthesis model and shows improved generalizability by training on a dataset with only 144 objects. 
Newbury \etal~\cite{Newbury23tro-graspReview} provides a nice review of the recent progress on end-to-end grasping models.  
The success of the end-to-end methods and the advent of vision-language models has enabled researchers to pose and address new questions such as task-oriented grasping\ralOmit{~\cite{Fang20ijrr-tog,Rashid23corl-lerfToGo}}, language-conditioned grasping\ralOmit{~\cite{Vuong24cvpr-languageGrasp,Rashid23corl-lerfToGo}}, dynamic object grasping\ralOmit{~\cite{Fang23tro-anyGrasp}}, and cross embodiment grasping\ralOmit{~\cite{Wei24icra-droGrasp}}.

It has been known in the grasping community that knowing object pose and shape, albeit via shape completion should enable better grasp synthesis. 
\ralAdd{Several works have proposed shape completion models to enable grasping, while others have argued shape completion as an auxiliary task to grasp synthesis. See~\cite{Newbury23tro-graspReview} for a nice review.}
\ralOmit{This has been investigated in many related works~\cite{Newbury23tro-graspReview}. 
Several works have proposed shape completion models to enable grasping~\cite{Gao21icra-shapeCompletionGrasp, Kiatos20tro-shapeCompletionGrasp, ChavanDafle22iros-shapeCompletionGrasp, Gualtieri21ral-shapeCompletionGrasp, Lundell19iros-shapeCompletionGrasp}. 
Another set of works have argued shape completion to be an auxiliary task to grasp synthesis~\cite{Jiang21rss-shapeCompletionAuxGrasp, Yan18icra-shapeCompletionAuxGrasp, Yang21icra-shapeCompletionAuxGrasp}. These works propose models that simultaneously perform shape completion and grasp synthesis.}
Recent examples include: 
Sen \etal~\cite{Sen23icra-scarpShapeCompletion} which proposes a method for shape completion, given partial-view point clouds, and demonstrates improved grasp synthesis in tabletop objects.
Wu \etal~\cite{Wu23rss-graspSuperquadrics} fits partial object point clouds with superquadrics and shows its utility in grasp synthesis. The work points out that the tri-symmetry property of superquadrics enables faster and simplified grasp generation.
%
% Mohammad \etal~\cite{Mohammad23icra-3DSgrasp} predicts missing geometry from single-view point cloud to synthesize grasps,
% while 
%
Chisar \etal~\cite{Chisari24ral-centerGrasp} proposes a simultaneous shape reconstruction and grasp synthesis.

\begin{figure*}[t]

\centering
\includegraphics[width=\linewidth]{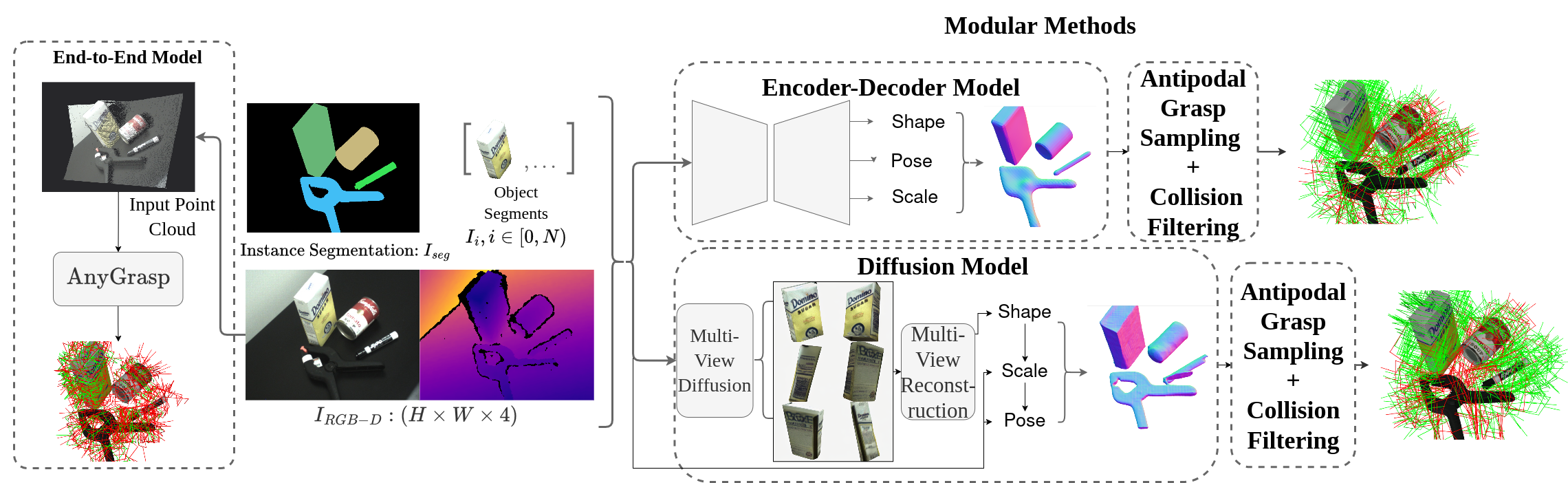}

\caption{Implementation of the four baselines: an end-to-end method that synthesizes grasps using a single-view RGB-D image, and three modular methods that first perform multi-object reconstruction from a single-view RGB-D image, followed by grasp synthesis using antipodal sampling and collision filtering. The three modular methods use two different ways of multi-object reconstruction: first, uses an encoder-decoder model, and the second, uses a diffusion model.}
\label{fig:arch}
\vspace{-3mm}
\end{figure*}

\section{Methods and Materials}
\label{sec:methods}

%\RT{
%
%\checkmark Describe the problem: that we wish to evaluate the grasp generation methods given single RGB-D image. 
%
%\checkmark Describe the pipeline.
%
%- Provide background on end-to-end grasp generation models.
%
%- Provide background on pose and shape estimation methods. Describe the two distinct methods: conditional distance fields and diffusion models. Discuss their workings.
%
%- \checkmark Describe what we aim to evaluate. 
%
%}

We consider the single-view grasp synthesis problem. 
In it, 
given a single-view RGB-D image $\MI$ of a table-top scene, 
the goal is to synthesize $n$ grasp poses $\MG\in \graspSpace$, where $\graspSpace$ denote the space of all grasp poses.
The parameterization of \graspSpace varies across methods~\cite{Sundermeyer21icra-contactGraspNet,Fang23tro-anyGrasp,Newbury23tro-graspReview}.
We scope our study to parallel-jaw grippers in which \graspSpace has a maximum of 7-DoF (rotation, translation, and optionally gripper width). 
%the goal is to synthesize $n$ grasp poses $\MG\in \SEthree^n$.
%the aim is to synthesize $n$ grasp poses $\MG\in \SEthree^n$. 
%
%A grasp pose is the three-dimensional pose of the robot hand and is represented as an element in \SEthree. 
%
%\RT{Gaurav -- I think we need a little more clarity here. In End-to-End Methods, we say it can be a 7 DoF grasp pose.}
%
The number $n$ of grasps synthesized is a free variable and depends on the input. 
We assume the camera intrinsics to be known.
%
%the camera intrinsics $\MK \in \mathbb{R}^{3\times3}$, the goal is to synthesize $n$ grasp poses $\MG\in \Real{n \times 4\times 4}$ on the scene in the camera coordinate frame. 
%%
%\RT{Can you add a bit more about the grasp pose?}
%
The problem is fundamentally ill-posed as the single-view RGB-D image does not provide complete information to enable grasp synthesis. %due to the partial nature of the input. 
However, 
today's state-of-the-art methods exploit large training data to learn enough priors and generalize (see related works in Section~\ref{sec:rel-grasp-synthesis}). %We analyze two types of methods in our work:
%therefore, a method must hallucinate or complete the unobserved regions and generate grasps corresponding to the full scene. Based on how this problem is tackled, we classify the methods to evaluate into two categories:
%
%\RT{Can motivate parallel gripper assumption as it is more general as stated in~\cite{Newbury23tro-graspReview}.}
%

%We implement and analyze three baselines across two grasp synthesis methods. The first, 
%
We implement, analyze, and compare four baselines across two types of grasp synthesis methods. The first, is the state-of-the-art \emph{end-to-end method} for grasp synthesis. The remaining three, are \emph{modular methods} that first estimate the object pose and shape of the objects in the scene, and then synthesize grasps using the antipodal grasp sampling in~\cite{Mahler17rss-dexnet2,Zhai22ral-da2dataset,Eppner19isrr-billionWaystoGrasp}. The three modular methods use two different ways of estimating the object pose and shape. %One uses the encoder-decoder model, CRISP~\cite{Shi25cvpr-CRISP}, and the other uses a diffusion model, SceneComplete (SC)~\cite{Agarwal24arxiv-sceneComplete}, for object pose and shape estimation of all objects in the scene.
%
%We use recent state-of-the-art models CRISP~\cite{Shi25cvpr-CRISP} and SceneComplete (SC)~\cite{Agarwal24arxiv-sceneComplete} for pose and shape estimation.
%
We do not train or fine-tune any of these models, but use publicly available pre-trained models, off-the-shelf.
Our choice of implementing these methods is driven by the fact that (i) they are now well established methods amongst researchers with many follow-up works that use them, (ii) they all show impressive performance and are very close to the state-of-the-art, and (iii) they have open-sourced their implementation and the pre-trained models are openly available. 
Figure~\ref{fig:arch} shows the inputs, outputs, and the relevant internal workings of the four baselines.

We next describe the two types of grasp synthesis methods: end-to-end (Section~\ref{sec:methods-end-to-end}) and modular (Section~\ref{sec:methods-modular}), and the four baselines. 
We the discuss our evaluation metrics (Section~\ref{sec:methods-eval}) and the experimental setup (Section~\ref{sec:methods-expt}). 
\subsection{End-to-End Methods for Grasp Synthesis}
\label{sec:methods-end-to-end}
End-to-end grasping models directly regress 6/7-DoF grasp poses (rotation, translation, and optionally gripper width) from a single-view RGB-D image or a partial point cloud. 
These models are trained end-to-end with grasp annotations, often with heavy augmentation across objects, clutter, and viewpoints. 
They aim to learn an implicit completion prior in the latent space while producing feasible grasps on partial scenes in a single forward pass. 
These methods have been favored for high accuracy, fast inference, generalizability to unknown objects and unseen environments, and robustness to occlusions.  
%However, the lack of an explicit shape-completion objective creates a non-interpretable setting where these models might become unreliable in settings with high partiality and domain shifts. \Madhav{What does high partiality mean?}
%

We implement \anyGrasp~\cite{Fang23tro-anyGrasp}, a state-of-the-art end-to-end grasp synthesis model. 
\anyGrasp is trained on a paired dataset of partial point clouds and 7-DoF grasp labels constructed by augmenting and extending large-scale grasp dataset GraspNet-1Billion~\cite{Fang20cvpr-graspnet1B} with additional objects and table-top scenes.
%Building on this base, 
\anyGrasp applies dense supervision (combining analytic and real-perception labels) across spatial and temporal domains to produce diverse, robust 7-DoF grasp candidates in a single forward pass~\cite{Fang23tro-anyGrasp}. 
% Please refer to their paper for more details.
We do not train or fine-tune \anyGrasp. We use the official \anyGrasp SDK released by the authors.\footnote{\url{https://github.com/graspnet/anygrasp_sdk}}
Figure~\ref{fig:arch} shows the inputs and outputs of the \anyGrasp model.

%Given a partial pointcloud $P$obtained from $\{K, I\}$, of a scene, this approach aims to learn a neural network to directly regress the grasp poses on the scene, in a feed-forward manner. This requires the model to be trained in a supervised manner on a paired dataset of table-top scenes with partial point clouds and corresponding grasp annotations~\cite{anygrasp}. In this setting, the model must learn to internally hallucinate the full scene in the underlying latent space as well as predict the grasp poses. For our experiments, we use AnyGrasp~\cite{anygrasp} as our model of choice.

\subsection{Modular Methods for Grasp Synthesis}
\label{sec:methods-modular}

We device and implement modular methods that first estimate the object pose and shape, of all objects in the scene, and then sample grasps. These methods differ in the way they estimate the object pose and shape. Given the object pose and shape we render a 3D object mesh by marching cubes~\cite{Lorensen87siggraph-marchingCubes}, and use the mesh to sample antipodal grasps. We use the sampling algorithm developed in~\cite{Mahler17rss-dexnet2,Zhai22ral-da2dataset,Eppner19isrr-billionWaystoGrasp} to sample antipodal grasps.
The estimated grasps, shapes, and poses are then fused together to obtain multi-object reconstruction and grasp estimates for the entire scene. The grasps that collide with other objects and structures in the scene are pruned out.   %
  
We implement two types of object pose and shape estimation methods, namely, encoder-decoder models and diffusion-based models. 
In encoder-decoder models, the encoder encodes the RGB-D image into an encoding that is either interpretable (\eg the normalized object coordinates~\cite{Wang19cvpr-nocs}) or abstract (\eg shape embedding~\cite{Wu23cvpr-mvCompressiveCoding,Hong24iclr-lrm,Shi25cvpr-CRISP,sam3d-objects}). The decoder decodes the encoding into an SDF or 3D Mesh of the object and outputs object pose. We implement CRISP~\cite{Shi25cvpr-CRISP} and SAM3D~\cite{sam3d-objects} for object pose and shape estimation, and refer to the corresponding modular methods as \crispGrasp and \samThreeDGrasp. 

Diffusion-based models, utilize pre-trained diffusion models that can be queried to generate a novel-view, given an RGB-D of an object and the relative camera pose. These diffusion models generate novel views and a separate network is trained to estimate the object's SDF. 
We use \sceneCompleteShort~\cite{Agarwal24arxiv-sceneComplete}, a recent multi-object scene reconstruction pipeline, that uses InstantMesh~\cite{Xu24arxiv-instantmesh} (a diffusion model for object shape estimation), and augments it with an in-painting model BrushNet~\cite{Ju24arxiv-brushNet} for robustness against occlusions and FoundationPose~\cite{Wen24cvpr-foundationPose} for pose estimation. 
\sceneCompleteShort extrapolates the zero-shot generalization capability of novel-view synthesis in InstantMesh to multi-object scene reconstruction. 
Figure~\ref{fig:arch} shows the workings of the modular grasp synthesis method that uses \sceneCompleteShort. 
We refer to this modular method as \scGrasp. %See Appendix~\ref{app:app-modular-methods} for more details. 

% \begin{figure}[h]
%     \centering
%     \includegraphics[width=0.48\linewidth,trim={50 100 50 100},clip]{figures/Cluttered_5.png}
%     \includegraphics[width=0.48\linewidth]{figures/arm_obj_1.jpg}
%     \caption{(a) Physics simulator scene. (b) Real-world experiment setup.}
%     \label{fig:expt-setup}
% \end{figure}

\subsection{Evaluation Metrics}
\label{sec:methods-eval}

We evaluate grasps using Grasp Collision Rate (GCR), Force Closure Failure Rate (FCFR), and Grasp Stability (GS). GCR measures the fraction of grasps invalidated by collisions, while FCFR checks whether the grasp can resist arbitrary external wrenches under friction cone constraints, together capturing grasp robustness and feasibility. GS checks how unstable the grasp is by seeing the object's motion during the pick. In our real-world experiments, a grasp is considered stable if the resulting pick exhibits minimal rotational deviation. %See Appendix~\ref{app:app-eval-metrics} for more details.

\subsection{Experimental Setup}
\label{sec:methods-expt}

We evaluate the grasp synthesis methods on physics simulator, real-world datasets, and real-world grasping experiments. We describe the setup of each of these environments. 
\subsubsection{Physics Simulator}
\label{sec:methods-expt-simulator}
\begin{figure}[h]
    \centering
    \includegraphics[width=0.48\linewidth,trim={50 100 50 100},clip]{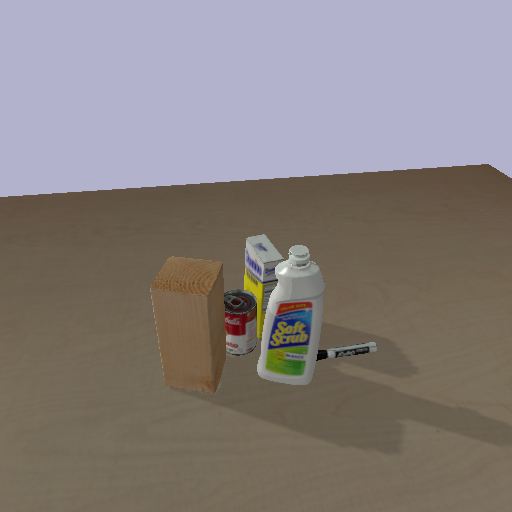}
    \includegraphics[width=0.48\linewidth]{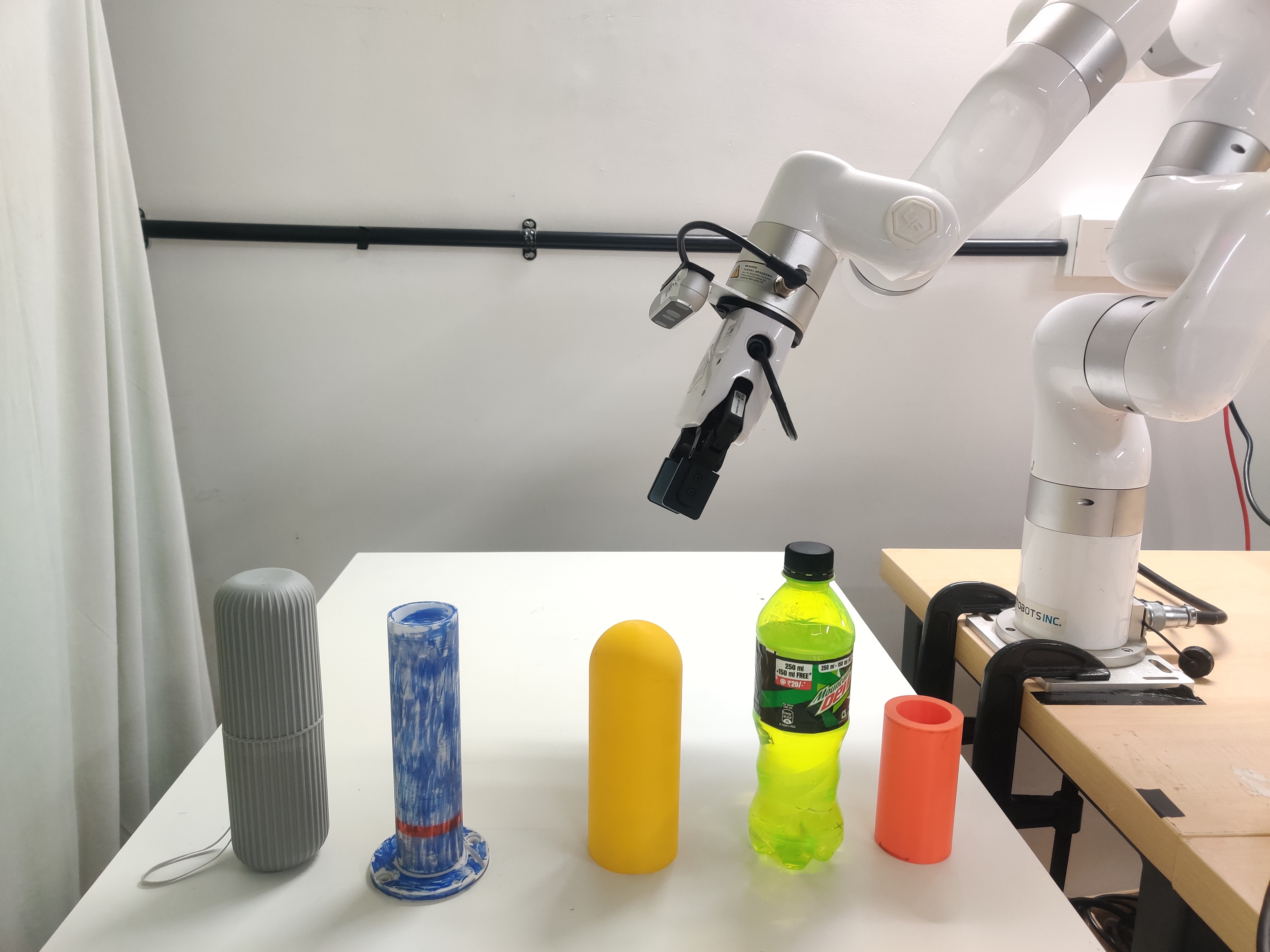}
    \caption{(a) Physics simulator scene. (b) Real-world experiment setup.}
    \label{fig:expt-setup}
\end{figure}

A physics simulator is the simplest way to analyze the performance of a grasp synthesis method without much cost overhead. 
We use the PyBullet simulator~\cite{pybullet24} and YCBV objects~\cite{Hodan20eccvw-BOPChallenge}.
We create multi-object table-top scenes by dropping  objects %\pavan{should we mention YCBV objects} 
near the center of a table. We run the PyBullet simulation %for six thousand steps and allow the objects to settle under gravity and collisions.  
%We run the simulation in PyBullet 
until the objects settle under gravity and collisions.
%We can phrase it this way instead of mentioning a fixed number of steps, as specifying the step count may not be meaningful.
%
%
We capture single-view RGB-D images from pre-determines camera orientations. We feed these images to our grasp synthesis methods. 
%The synthesized grasps are evaluated using Grasp Collision Rate (GCR) and Force Closure Failure Rate (FCFR) metrics. 
%
%
Figure~\ref{fig:expt-setup}(a) shows a rendered image of our scene in the Physics simulator.

\subsubsection{Real-World Datasets}
\label{sec:methods-expt-datasets}
It is hard to find a dataset that has both pose and shape annotations as well as grasp annotations. We relied on real-world datasets that have good pose and shape annotations.
We use YCBV and NOCS datasets~\cite{Wang19cvpr-nocs, Hodan20eccvw-BOPChallenge} for evaluating our baselines. In particular, we only used the test sets in the two datasets. 
The test sets contain several images of a multi-object scene, taken from a video stream. The test set contains multiple scenes. As two consecutive images share a large overlap we sampled two random images for each scene and collected them for our evaluation. %This was done mainly to save time. We did not observe much variation when we ran evaluations for all the views. 

\subsubsection{Real-World Grasping Experiments}
\label{sec:methods-expt-real}
We conduct real-world experiments with Xarm7 manipulator. 
The Xarm7 is equipped with parallel-jaw grippers, which move to the grasp pose and close from the maximum opening width.
We use single-view RGB-D images captured by an Intel RealSense D455 camera placed on the Xarm7. 
%\RT{Pavan: do we have an image of this?} 
%\textcolor{teal}{Pavan: No, but we can take image of xarm + d455 if its needed}
%
We conduct experiment with hand-designed objects primarily to ensure safety and durability of our Xarm7 manipulator.
%\pavan{I don’t see the connection between object selection and safety—could you clarify?}  RT -- we are using 3D printed objects and not real objects. This is mainly to ensure safety of our Xarm7. It is a good thing to mention otherwise reviewers will complain that we are not thorough in our real world experiments. 

%\RT{Pavan: can you confirm?} 
%\textcolor{teal}{Pavan: The objects included Mountain Dew bottle, one toothbrush case, and three 3D-printed objects. They were not chosen to ensure durability of the arm, so this statement should be revised.}
The objects include a soda bottle, a grey case, tall blue jar, and two 3D-printed objects: yellow stopper and orange roll. %\RT{Pavan - (i) where are the results for toothbrush case? (ii) are all 3D printed objects cylinders?} 
%\pavan{Correction: The objects included a soda bottle, a grey case and three 3D-printed objects: tall blue jar, yellow stopper and orange roll.}
%
To ensure safety of the Xarm7 the grasps that have too low approach angle vis-a-vis the table are filtered out. The grasps below a minimum clearance distance from the table were also filtered to avoid collisions. %\RT{Pavan: what does this mean "Gripper height $>$ object-dependent threshold"?}
%\textcolor{teal}{Pavan: It means the gripper must stay above a minimum clearance from the table to avoid collisions, which can be generalized as a minimum height constraint rather than an object-dependent threshold}
%
These filtered grasps are not considered for evaluation.
In our real-world experiments, a grasp is considered \emph{successful} if our arm can lift the object; otherwise, it is counted as a \emph{failure}. The grasps are further evaluated based on the Grasp Stability (GS) criteria.

\section{Experimental Analysis}
%\label{sec:expt}

% The main goal of our experimental analysis is to answer the question raised in the abstract. Towards this, we implement and compare the three baselines: \anyGrasp, \crispGrasp, \scGrasp. 
% %
% We conduct experiments in both a physics simulator environment (Section~\ref{sec:expt-controlled}) and real-world environments (Section~\ref{sec:expo-datasets}).
% %
% We investigate and analyze the impact of object size and scene clutter (Section~\ref{sec:expt-size-and-clutter}).  
% %
% We also implement and analyze two more baselines in these experiments: \anyGraspFilterCRISP and \anyGraspFilterSC. These baselines use the estimated object pose and shape via \crisp and \sceneCompleteShort to filter out colliding grasps. 
% This allows us to investigate whether such a trivial augmentation of object pose and shape estimation into the end-to-end grasp synthesis method results in any significant advantages. 
% %
% %
% Finally, we dive deeper into the two modular approaches and
% %We 
% analyze the failure modes (Section~\ref{sec:expt-failure}) and run-times (Section~\ref{sec:expt-time}) of the two modular methods.
% %
% We augment the modular methods with vision-language models to yield task-oriented grasps from just sigle-view RGB-D image as input (Section~\ref{sec:expt-tog}). 

%\RT{
%
%- Change figure labels: CRISP -to- CRISP Grasp, SC -to- SC Grasp, Anygrasp -to- AnyGrasp.
%}

\subsection{Physics Simulator Experiments}
\label{sec:expt-controlled}
\begin{figure}
	\centering
    \includegraphics[width=0.8\linewidth,trim=0 10 0 232, clip]{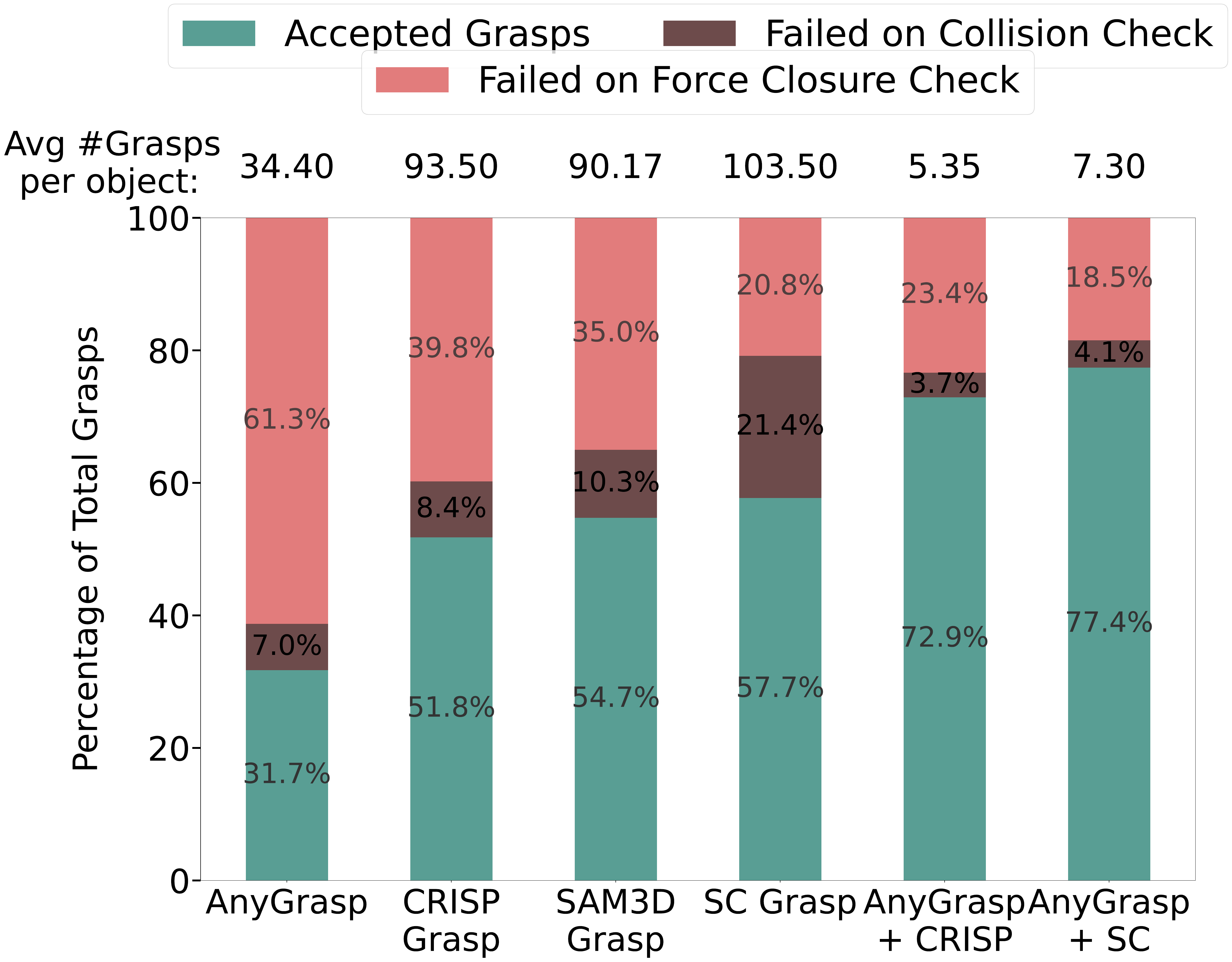}
    % Legend
    % \vspace{0.5em}
%    \small{
%    \begin{tabular}{ccc}
%        \textcolor{myred}{\rule{0.7em}{0.7em}} Accepted Grasps &
%        \textcolor{myblue}{\rule{0.7em}{0.7em}} Failed on Collision Check
%    \end{tabular}
%    \begin{tabular}{c}
%        \textcolor{mygreen}{\rule{0.7em}{0.7em}} Failed on Force Closure Check
%    \end{tabular}
%    }
    \vspace{-2mm}
	\caption{\textbf{Physics Simulator Experiments:} Plots the grasp success rate (\textcolor{mygreen}{\rule{0.7em}{0.7em}}), collision rate (\textcolor{mybrown}{\rule{0.7em}{0.7em}}), and force closure rate (\textcolor{myred}{\rule{0.7em}{0.7em}}) for the five baseline methods: \anyGrasp, \crispGrasp, \samThreeDGrasp, \scGrasp, \anyGraspFilterCRISP, and \anyGraspFilterSC. Also shown is the average number of grasps synthesized per object for each baseline method.}
	\label{fig:PybulletSimulation}
    \vspace{-5mm}
\end{figure}

We observe that grasp success rate for the modular methods (\ie \crispGrasp, \samThreeDGrasp, and \scGrasp) is 1.6 to 2 times higher than the end-to-end method (\ie \anyGrasp). See Figure~\ref{fig:PybulletSimulation}.
%
% Figure~\ref{fig:PybulletSimulation} plots the grasp success rate (GSR, \textcolor{mygreen}{\rule{0.7em}{0.7em}}), grasp collision rate (GCR, \textcolor{mybrown}{\rule{0.7em}{0.7em}}), and the force closure failure rate (FCFR, \textcolor{myred}{\rule{0.7em}{0.7em}}), in percentages, for the three baseline methods. 
Synthesized grasps fail mainly because they do not satisfy the force closure constraints. 
The end-to-end method yields a very high force closure failure rate of $61.3\%$.
Accurate pose and shape estimation, coupled with antipodal sampling that explicitly accounts for force closure, yields this result. The end-to-end method relies on generalizability to unseen environments and objects, and fails to match in performance.
%

%
%
% Figure~\ref{fig:PybulletSimulation}  
% %shows the average number of grasps synthesized by a method per object 
% % \pavan{For \anyGrasp, Kushal applied a thresholding step to remove grasps lying far from the object. 
% % The remaining grasps were then evaluated, and their counts are reflected in the plot. 
% % Hence, the reported grasp counts may not directly represent the raw grasps produced by the model, 
% % since some were discarded during this filtering.}
% % \red{shows the total number of grasps synthesized by a method across all objects}; 
% shows the average number of grasps synthesized by a method (per object) across all objects; see top of the figure. 
%

In Figure~\ref{fig:PybulletSimulation}, we also plot two more methods, \anyGraspFilterCRISP and \anyGraspFilterSC, that use pose and shape estimation to filter out grasps synthesized by the end-to-end method. We filter out the grasps that collide with the reconstruction or do not satisfy force closure. While the filtering drastically reduces the average number of grasps per object (top of Figure~\ref{fig:PybulletSimulation}), we observe that it significantly increases the grasp success rate. This shows that the filtering of grasps by the pose and shape estimates retains the successful grasps.

We note that the modular methods are able to synthesize many more grasps as per our specification; we cap this to approximately $100$ in our experiments. 
% See the top row of Figure~\ref{fig:PybulletSimulation}. 
%
The end-to-end method, on the other hand, generates far fewer grasps per object on average, leaving very few successful grasps to grasp an object. 
Figure~1(a) also visualizes the synthesized grasps for an input instance. Full coverage produced by the antipodal sampling procedure is a major advantage of the modular method that the end-to-end method lacks.

\begin{figure*}

\centering
\includegraphics[width=.4\linewidth, trim= 0 0 0 232, clip]{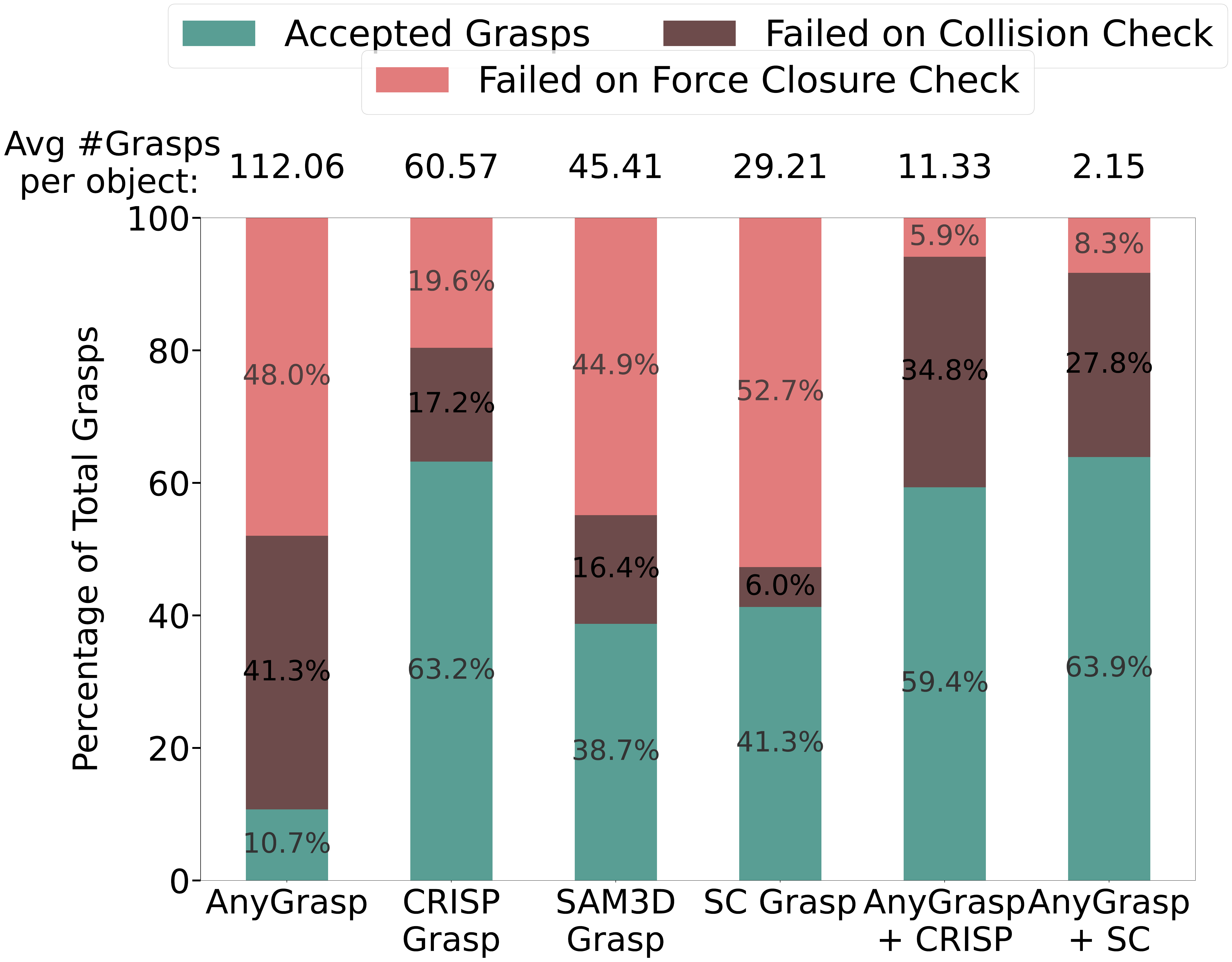}
\hspace{5mm}
\includegraphics[width=.4\linewidth, trim= 0 0 0 232, clip]{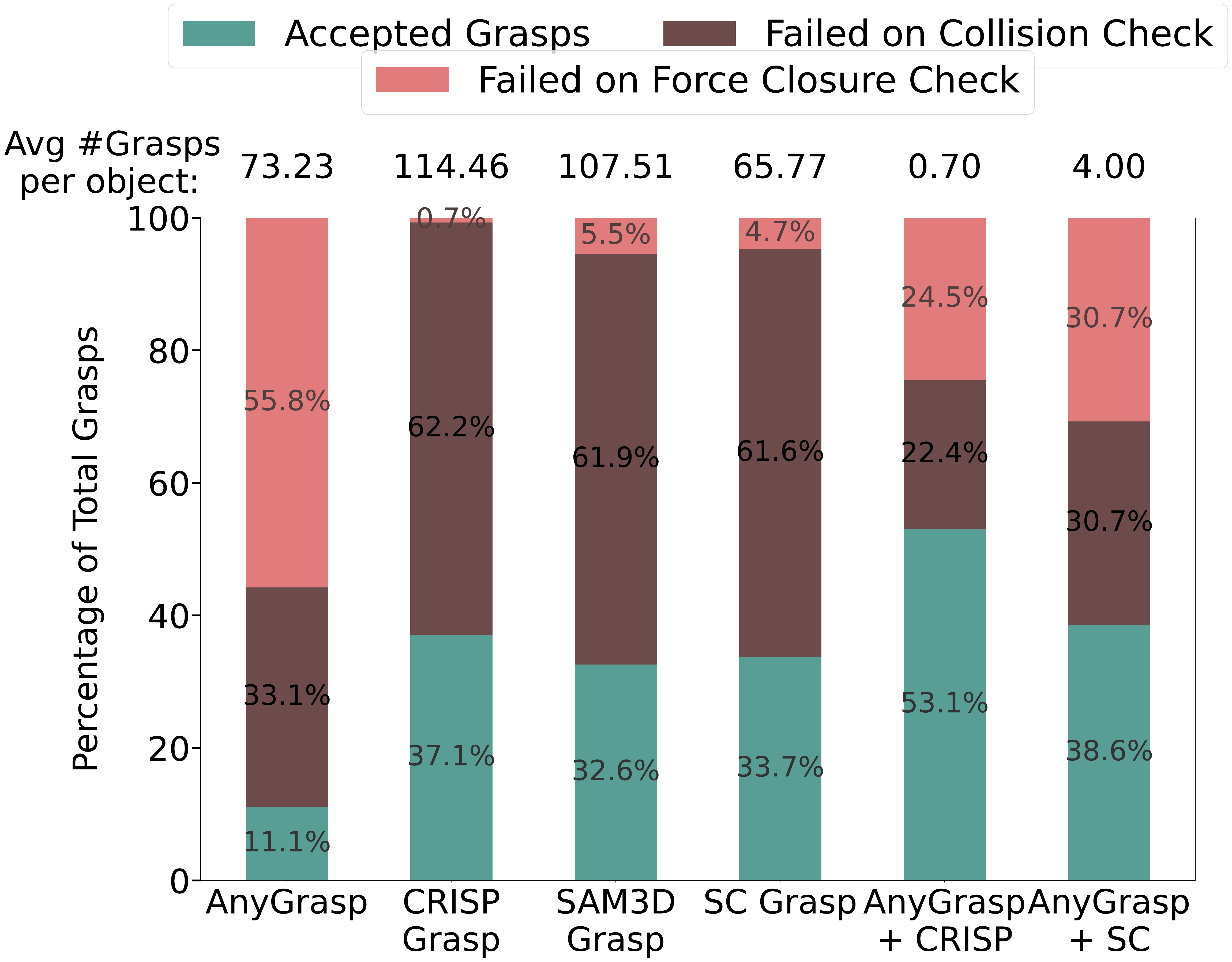}
\vspace{-1mm}
\caption{\textbf{Real-World Dataset Experiments:} The two figures plot the grasp success rate (\textcolor{mygreen}{\rule{0.7em}{0.7em}}), collision rate (\textcolor{mybrown}{\rule{0.7em}{0.7em}}), and force closure rate (\textcolor{myred}{\rule{0.7em}{0.7em}}) for the five baseline methods: \anyGrasp, \crispGrasp, \samThreeDGrasp, \scGrasp, \anyGraspFilterCRISP, and \anyGraspFilterSC. Also shown is the average number of grasps synthesized per object for each baseline method. \textbf{Left} plots for the YCBV test dataset. \textbf{Right} plots for the NOCS test dataset.}
\vspace{-2mm}
\label{fig:ycbv-nocs}

\end{figure*}

\subsection{Real-World Dataset Experiments}
\label{sec:expo-datasets}

After validating our hypothesis in a controlled, physics simulator experiments, we study the methods on real-world datasets. 
We consider YCBV and NOCS dataset, as they come with annotated poses and 3D object meshes. 
% We evaluate all baselines on these datasets and present the results in Figure~\ref{fig:main}. We made a couple of modifications to the evaluation to account for multiple objects in a scene. Firstly, for our Modular methods like \scGrasp and \crispGrasp we use the reconstructions of the other objects in the scene, to prune out grasps that would collide with other objects in the scene. Similarly while evaluation we check collisions of the grasps with all objects in the scene. Allowing us to test the validity of the grasp in the scene, rather than as a standalone. Secondly, since \anyGrasp does not map the grasps to the objects in the scene, we use the segmentation masks with a distance threshold, to assign the grasps to each object.
% % \pavan{ Kushal: can you add point for 1)antipodal+multiple objects, which describes that you have pruned antipodal  grasps for a mesh which collided with other meshes, this is done for all objects ,then evaluation on GT mesh is done \\
% % 2)add a point about how thresholding is done in anygrasp Evaluation at appropriate section
% % } \\
%
The observations we made in Section~\ref{sec:expt-controlled} largely hold for the real-world datasets as well. 
The modular methods significantly outperform the end-to-end method in real-world datasets as well. 
See Figure~\ref{fig:ycbv-nocs}.
The end-to-end method achieves a grasp success rate of only about $10-11\%$, where as this number is much higher for the modular methods.

A key observation is that \scGrasp and \samThreeDGrasp performs worse than \crispGrasp for the YCBV dataset (left of Figure~\ref{fig:ycbv-nocs}), which is in contrast to the controlled experiment. This is primarily because the YCBV scenes contain occlusions which impacts \samThreeD and the inpainting model of \sceneCompleteShort (see Section~\ref{sec:expt-size-and-clutter} where we investigate the impact of occlusions on the grasp success rate). %\red{Follow up in clutter experiments.}
For NOCS (right of Figure~\ref{fig:ycbv-nocs}), however, we do not observe a much difference between the grasp success rates of the three modular methods. This is because the NOCS scenes, although multi-object, have less occlusions than the YCBV scenes. 
%
% In Figure~\ref{fig:ycbv-nocs}, we note that there is an increase in the grasp collision rate as compared to the physics simulator experiments (\ie fig.~\ref{fig:PybulletSimulation}). 
% This is because the real-world datasets contain multiple objects in the scenes and an error in reconstruction of any one of the objects can cascade into extra collisions of the synthesized grasps with other objects. %leads to the extra collisions with the ground truth of that object. 
%
We also find that the two baselines \anyGraspFilterCRISP and \anyGraspFilterSC, which use the estimated pose and shape to filter out unsuccessful grasps, show
gains diminished vis-a-vis the end-to-end method, and  
only marginal improvement over \crispGrasp and \scGrasp, respectively. 

\subsection{Real-World Grasping Experiments}
% \RT{
% We do real-world grasping experiments to show 

% - Our analysis in controlled setting and on the real-world dataset hold true for the real-world. We report the following measures. 
% }

% \textbf{Videos and more details:} \href{https://docs.google.com/document/d/1DcoLRgjnM1R9PwDoKyV-mte581XknQ_I3EJ1zsmHpqM/edit?pli=1&tab=t.0}{Google Doc Link} 

% {
% \textit{Note on Grasp Stability:} Some AnyGrasp grasps showed post-lift instability, likely due to asymmetrically placed contact points—a result of relying on partial point cloud input. \\
% Video: \href{https://iiithydresearch-my.sharepoint.com/personal/pavan_karke_research_iiit_ac_in/_layouts/15/stream.aspx?id=%2Fpersonal%2Fpavan_karke%5Fresearch%5Fiiit%5Fac%5Fin%2FDocuments%2FAttachments%2FVID%5F20250708%5F155216%2Emp4}{Watch Video}
% }

% {
% \textit{Faliure Case:} AnyGrasp, trained end-to-end on partial point clouds, may miss global object features like diameter. Our method, using antipodal sampling on full mesh reconstructions, enables better-aligned contacts and more stable grasps. \\
% Video: \href{https://iiithydresearch-my.sharepoint.com/personal/pavan_karke_research_iiit_ac_in/_layouts/15/stream.aspx?id=%2Fpersonal%2Fpavan%5Fkarke%5Fresearch%5Fiiit%5Fac%5Fin%2FDocuments%2FAttachments%2FVID%5F20250707%5F164430%201%2Emp4&ga=1&referrer=StreamWebApp.Web&referrerScenario=AddressBarCopied.view.2e0b4dc6-be13-44ff-b00e-4e5d1ba4a901}{Watch Video}
% }

In order to validate our hypothesis further, we carry out trails with manipulator arm grasping and lifting several objects we have in our lab. We run trials with five different objects with different shapes, texture, and material properties. We run 7-10 trials for each object and observe the grasping. We only run \scGrasp in these experiments as \crispGrasp require re-training the model for new objects.  

\begin{table}[h]
\centering
\vspace{-3mm}
\caption{Successful and Stable grasps per object}
\begin{tabular}{|l|c|c|c|c|}
\hline
\textbf{Object} & \textbf{Trials} & \scGrasp & \multicolumn{2}{c|}{\anyGrasp} \\
& & Succ. \& Stable & Succ. & Stable \\
\hline
yellow stopper     & 10 & 100\% & 100\% & 100\% \\
grey case          & 10 & 100\% & 100\% &  80\% \\
orange roll        &  7 & 100\% &  43\% &  43\% \\
tall blue jar      & 10 & 100\% & 100\% &  60\% \\
soda bottle        & 10 & 100\% &  90\% &  90\% \\
\hline
\end{tabular}
\label{tab:grasp_results}
\end{table}
Surprisingly, the modular method \scGrasp succeeds and produces stable grasps $100\%$ of the times. On the other hand, the end-to-end method shows many failures. Table~\ref{tab:grasp_results} reports the successful and stable grasps in all our trials. 
The success rates in real-world experiments are higher compared to the controlled and dataset experiments. This is because in the real-world experiments we use a soft contact patch to stabilize the grasps. 
We analyze each failure instance. 
The end-to-end method produces unstable grasps for the blue jar in four out of ten instances. These exhibit post-lift rotations. These failures arise from asymmetric contacts due to partial point clouds, resulting in torque imbalance. In contrast, the antipodal sampling in the modular method consistently produces symmetric contacts and stable orientations. 
The orange roll has a low friction coefficient compared to other objects. A grasp on the orange roll fails even if the grasp location has even small deviation from the center grasp pose. The object slides out of the grasp in all the four our of seven failing instances. 
The soda bottle being translucent impacts \anyGrasp more than the modular methods which are able to successfully estimate its pose and shape using RGB. The noisy point cloud results in failure of \anyGrasp in this instance.   
\subsection{Cluttered Scene and Occlusions}
\label{sec:expt-size-and-clutter}
We investigate how the cluttered scene and occlusion affect the performance of the baseline and our hypothesis. 
We generate scenes with $1$, $5$, and $10$ objects in the physics simulator (Section~\ref{sec:methods-expt-simulator}). We compare the performance of the three baselines: \anyGrasp, \crispGrasp, and \scGrasp. 

The results are presented in Figure~\ref{fig:clutter}. 
We observe a drop in performance as the clutter increases across all baselines. 
The performance of \anyGrasp drops marginally from a very low base (\ie 1 Object scene), while that of \scGrasp sees the largest drop from a very high base. 
This is because \sceneCompleteShort uses an in-painting model to overcome the issue of occlusions and the in-painting model fails to generalize as expected. 

We notice a higher grasp collision rate for \anyGrasp than for the modular methods. This is because (i) the end-to-end method does not have the exact estimate of the object pose and shape, and (ii) in modular methods we are able to filter grasps that collide with other objects, which is not possible in the end-to-end method. 
We conclude that even in cluttered environments (10 objects packed closely in a scene), \crispGrasp outperforms the end-to-end method. %\anyGrasp, reinforcing the improved performance of the modular method over the end-to-end method. %We claim that using multiple inpainting seeds for \scGrasp and picking the best one would allow \scGrasp to outperform end-to-end methods in high clutter environments as well.

\subsection{Failure Analysis}
\label{sec:expt-failure}

We analyze the failure modes of the two modular methods: the encoder-decoder based \crispGrasp, \samThreeDGrasp, and the diffusion-model based \scGrasp. We ask when the modular methods result in a failing grasp, is it because of the incorrect shape, scale, or pose. We conduct this experiment in the physics simulation environment. We use a single object per scene to obviate the impact of other factors like clutter and occlusion. We examine and manually annotate the failure modes on each grasp failure instance using a set procedure. %; see Appendix~\ref{app:app-failure-analysis} for details.
Our annotators visualize the ground-truth object and its corresponding reconstruction. If the shape is incorrect, a shape error is declared. If shape is deemed correct, then the annotator checks for scale error. If the scale too is deemed correct, the annotator checks for pose error. 

% We divided our failure cases into three broad categoires,
% failure of Shape, Scale and Pose, SceneComplete internally
% follows this hierarchy when estimating the reconstruction and it
% is justified since an error in Shape (where you may consturct an
% apple instead of a bowl, ultimately affects Scale and Pose of the
% reconstruction as well, similarly an incorrect scale could lead
% to errors in the pose estimation. The overall classification has
% been subjective, where we visualize the ground truth object and
% it’s corresponding reconstruction at the same time to identify
% the primary causes of failure and pick categories based on that.
% For example, if there is a massive error in the shape, we allow
% for minor errors in the pose and the scale, but a similar error
% with a correct shape will be considered as a failure.
%
% We explored using an automated pipeline with custom thresholds to categorize the failures, we observed that certain failing grasps were due to errors in multiple aspects, and the errors were interdependent (\eg an error in scale, would ultimately impact the pose estimation as well). 
%We decided to manually annotate the failures, using our understanding of the approaches to identify the primary causes of the failure. The results are presented in Figure \ref{fig:failure-analysis}. We observe the following.

\begin{figure}
% \vspace{-2mm}
\centering
\includegraphics[width=0.32\linewidth, trim= 10 60 10 50, clip]{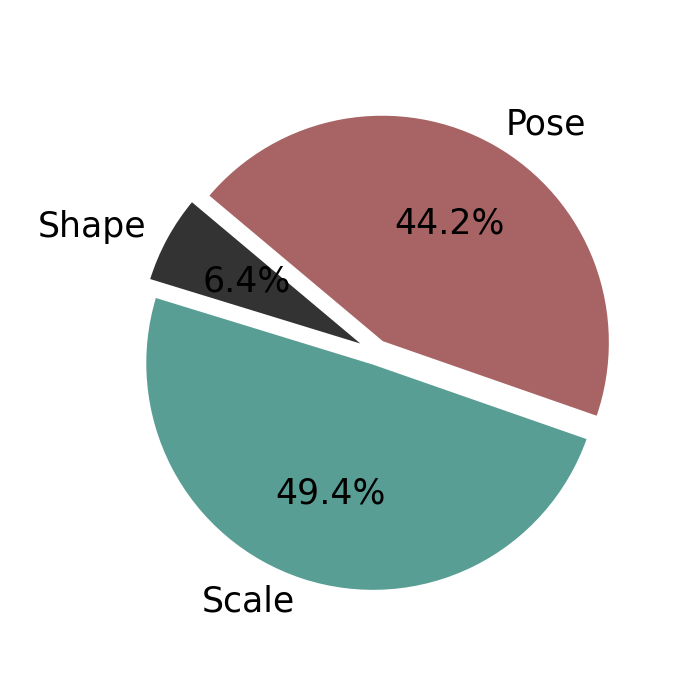}
\includegraphics[width=0.32\linewidth, trim= 5 60 5 50, clip]{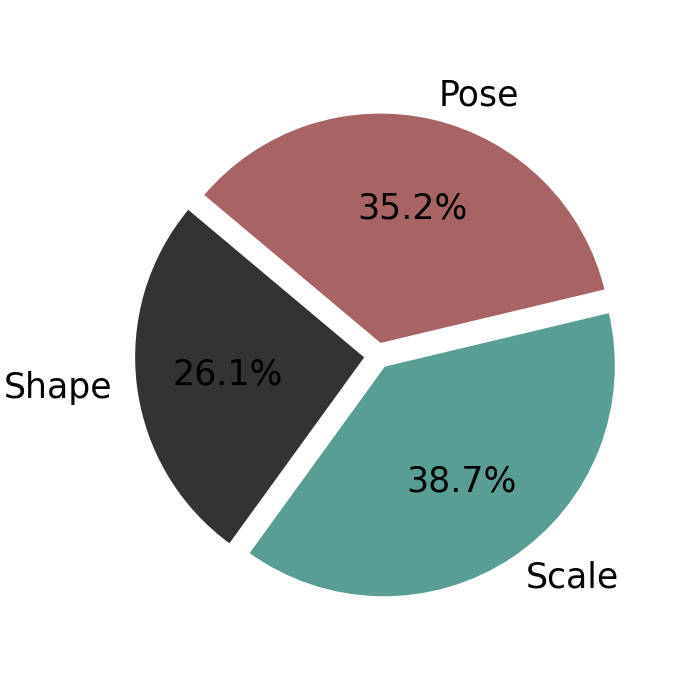} 
\includegraphics[width=0.32\linewidth, trim= 10 60 0 50, clip]{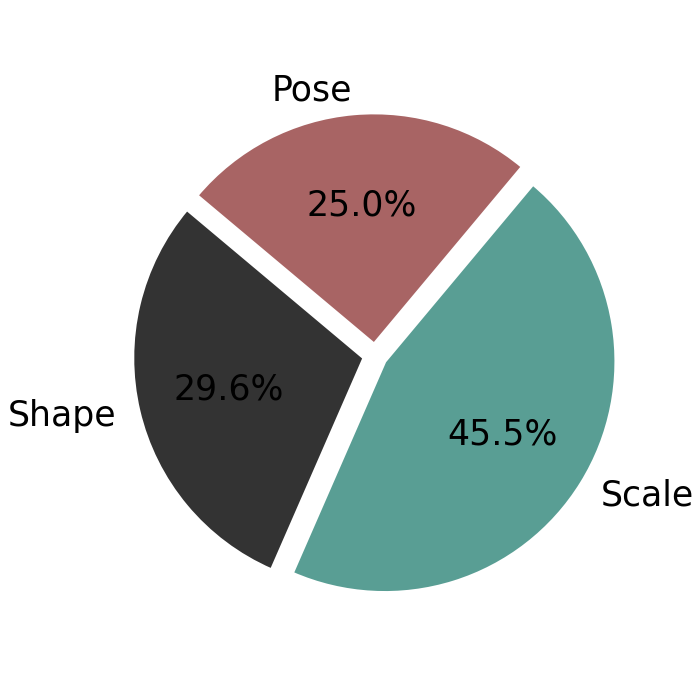} 
\caption{Failure models of \crispGrasp (left), \samThreeDGrasp (middle), and \scGrasp (right).}
\vspace{-4mm}
\label{fig:failure-analysis}

\end{figure}

\begin{figure*}

\centering
\includegraphics[width=.32\linewidth, trim= 0 0 0 232, clip]{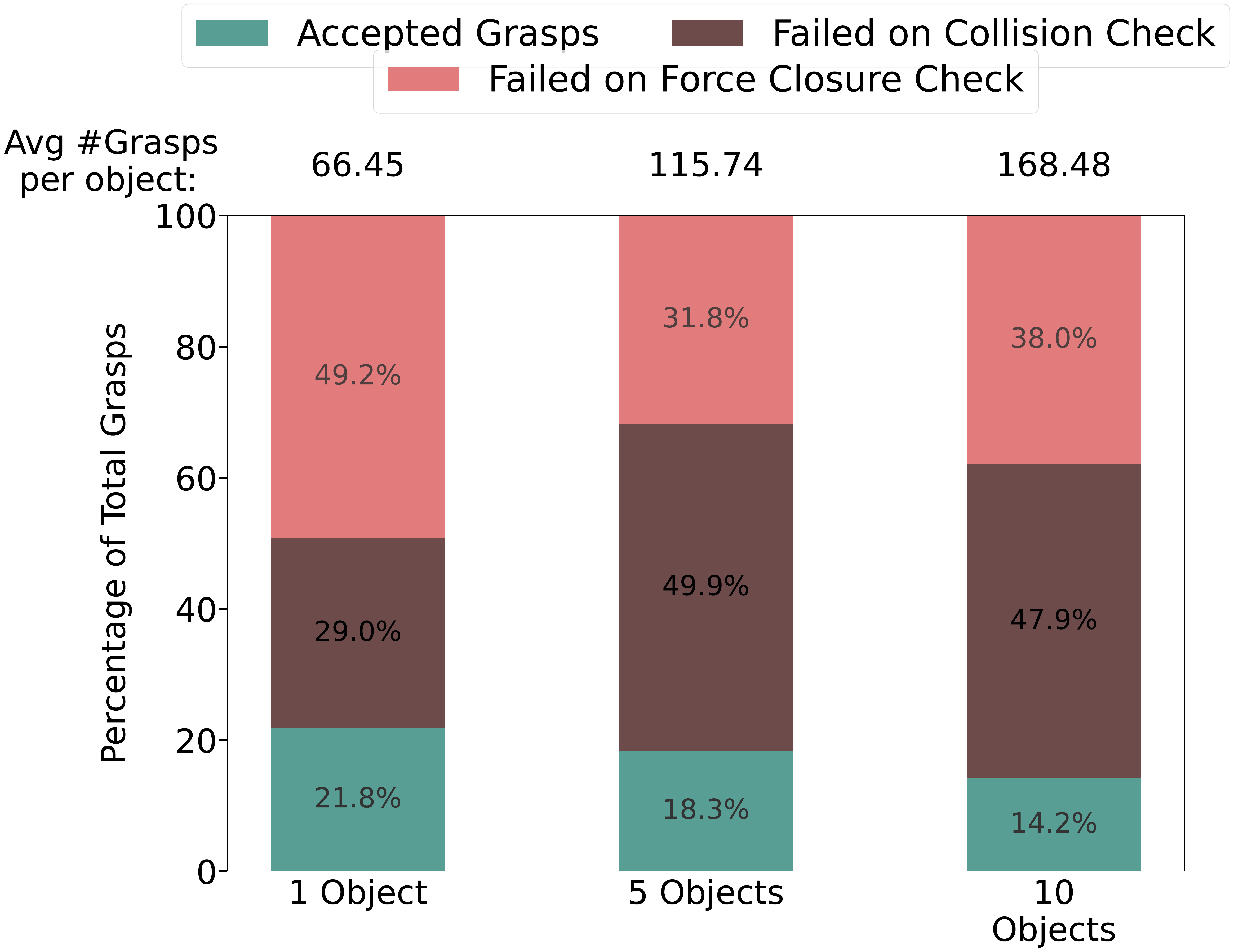}
\includegraphics[width=.32\linewidth, trim= 0 0 0 232, clip]{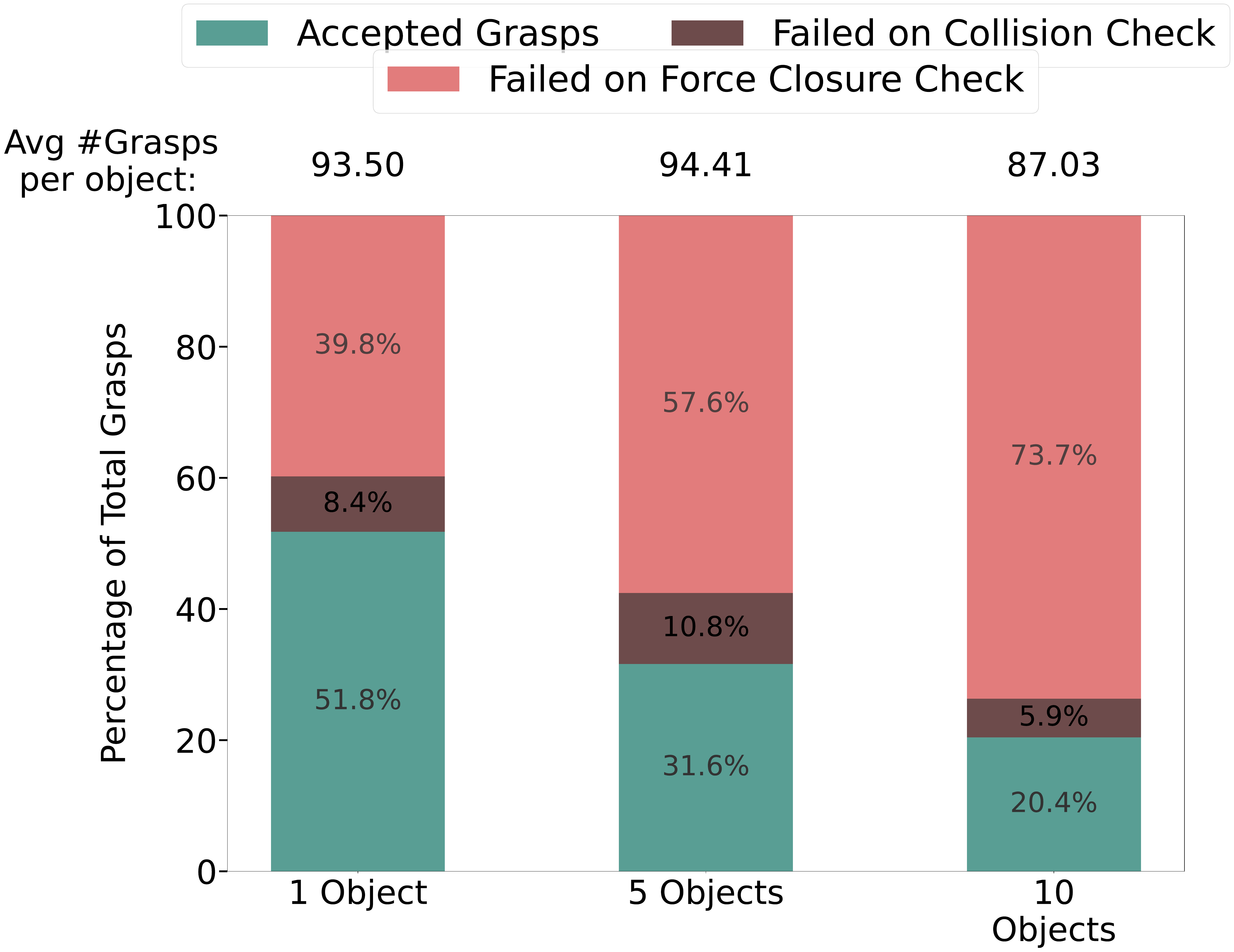}
\includegraphics[width=.32\linewidth, trim= 0 0 0 232, clip]{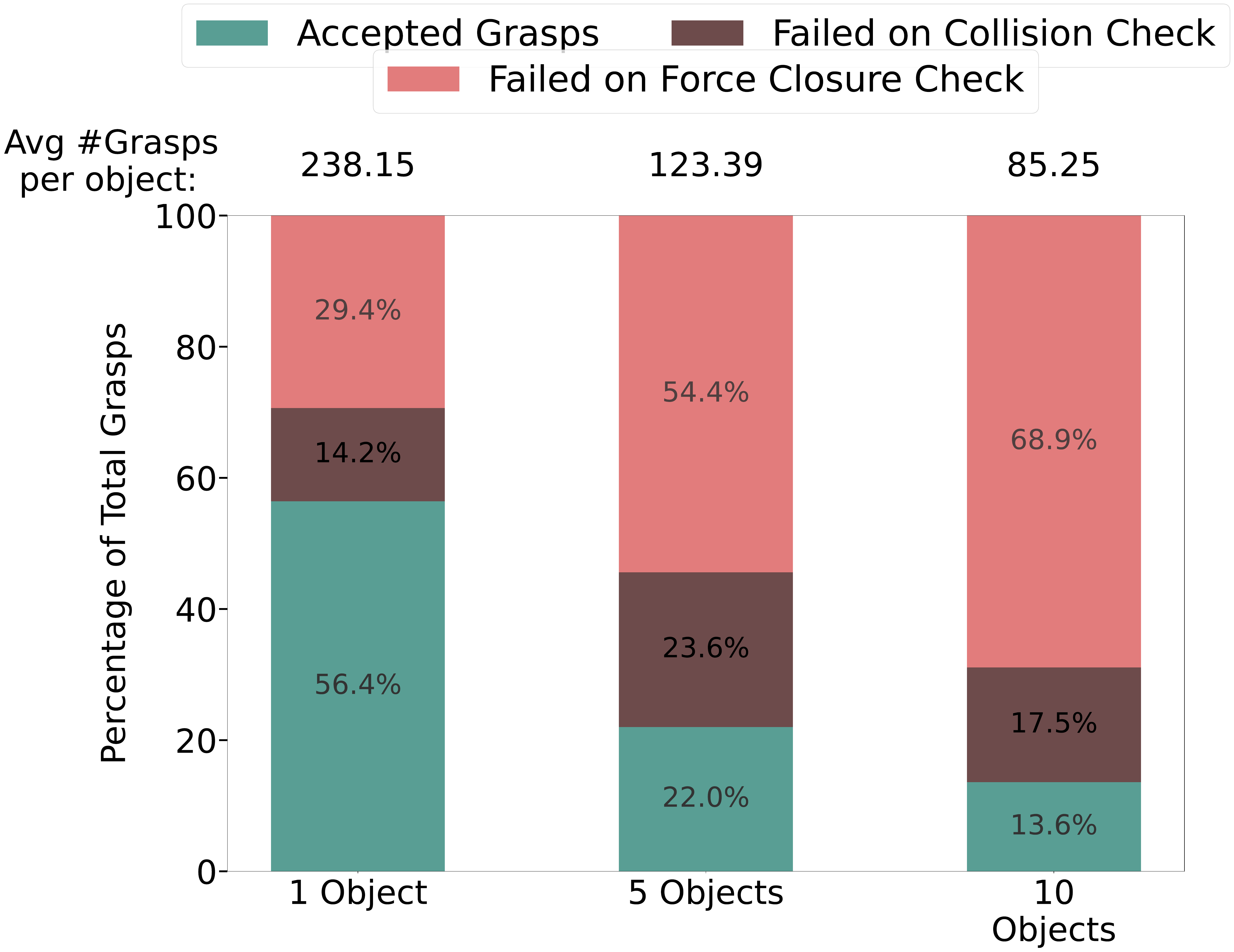}

\caption{\textbf{Grasp Generation in Cluttered Environments:} The three figures plot the grasp success rate (\textcolor{mygreen}{\rule{0.7em}{0.7em}}), collision rate (\textcolor{mybrown}{\rule{0.7em}{0.7em}}), and force closure rate (\textcolor{myred}{\rule{0.7em}{0.7em}}) for the three types of clutter: 1 object, 5 objects and 10 objects. Also shown is the average number of grasps synthesized per object for each type of clutter. \textbf{Left} plots for \anyGrasp baseline. \textbf{Middle} plots for \crispGrasp baseline. \textbf{Right} plots for the \scGrasp baseline.}
% \vspace{-3mm}
\label{fig:clutter}

\end{figure*}

We observe that incorrect scale estimation is a major cause of failure in modular methods. About $38$-$50\%$ of the failed grasps are attributed to incorrect scale estimates in our experiments (see Figure~\ref{fig:failure-analysis}).
We observe that \crispGrasp failure is rarely ($6.4\%$) attributable to shape estimation errors. This shows that the shape error in \crisp is not significant enough to cause a grasp failure. 
\samThreeDGrasp and \scGrasp, on the other hand, shows approximately equal chance that the failed grasp is a result of a pose and shape error ($25$-$35\%$). 
This makes intuitive sense as \crisp is designed to estimate shape of a closed set of objects unlike \sceneCompleteShort and \samThreeD.

\begin{figure*}
\centering
\includegraphics[width=0.85\linewidth]{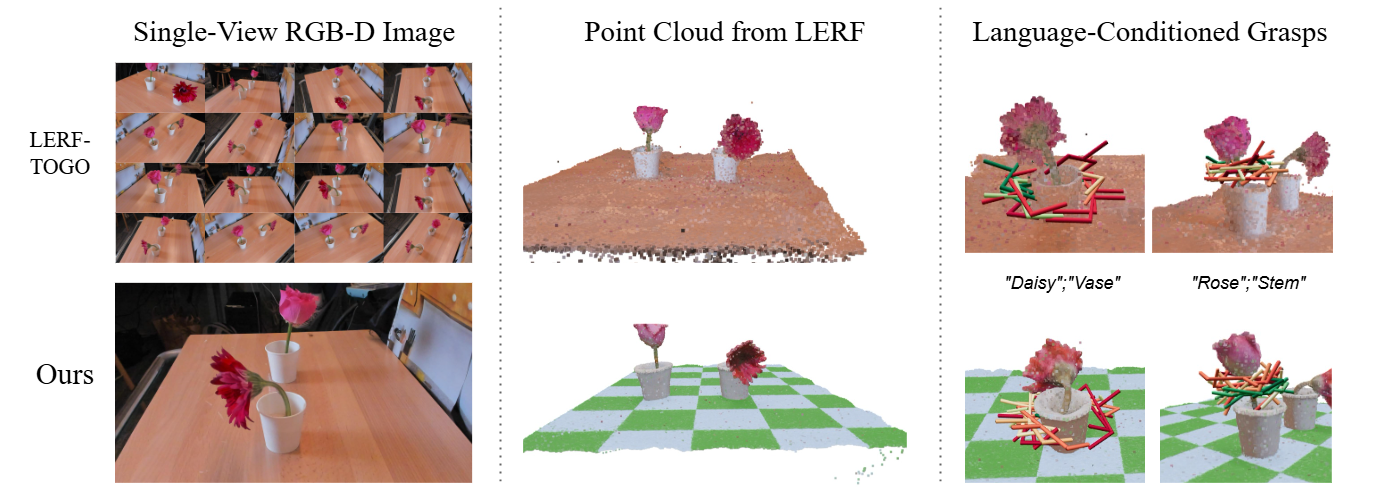}
\caption{Comparison of \lerfToGo and our single-view pipeline. Unlike \lerfToGo, which relies on multi-view input, our approach reconstructs a digital twin from a single RGB-D image. The resulting point cloud preserves geometry and texture, enabling language-conditioned grasps that are both stable and semantically meaningful.}
\vspace{-3mm}
\label{fig:lerf}
\end{figure*}

\subsection{Run-Time Analysis}
\label{sec:expt-time}

We analyze the run-time requirements for each of the baselines. 
We run all the methods on NVIDIA RTX 6000 Ada Generation GPU. 
We observe that the end-to-end method has the fastest run-time of $0.33$s. The diffusion-based \scGrasp takes longer than the encoder-decoder based methods \crispGrasp and \samThreeDGrasp %; both of which take considerably longer, clocking at $188.18$s and $38.60$s, respectively. 
(see Table~\ref{tab:compute}).
The run-time for \crispGrasp and \samThreeDGrasp is dominated by antipodal grasp sampling and filtering operation whereas the pose and shape estimation is relatively fast, clocking at $13.78$s and $8.68$s, respectively.
% Antipodal sampling takes a large overhead for both the modular methods ($30$-$40$s), making them unbeatable vis-a-vis the end-to-end method in run-time. 
%
The run-time in diffusion-based \scGrasp is dominated by the pose and shape estimation module. 
% While diffusion models in \scGrasp dominate the run-time, the pose and shape estimation in \crispGrasp is dominated by an optimization-based refinement step. 
%
This shows that for the modular methods to outperform the end-to-end method, we will need progress on both faster pose and shape estimation and faster antipodal sampling.

\vspace*{0.75em}

\begin{table}[h!]
\vspace{-5mm}
\caption{Run-time Analysis}
\vspace{-3mm}
\begin{center}
\begin{tabular}{cccc} 
 \hline
 Method & Pose+Shape Est. & Grasp Sampling & Total \\ 
\hline
 \anyGrasp & - & 0.33s & 0.33s
\\ 
 \crispGrasp & 13.78s & 31.51s & 45.29s
 \\ 
\samThreeDGrasp  & 8.68s  & 37.09s & 45.77s
\\ 
 \scGrasp & 65.20s & 39.71s & 104.91s

\\

%  \anyGraspFilterCRISP &
 
% 7.09

% & 
% 0.61

%  & 
 
% 7.70
%  \\ 
%  \anyGraspFilterSC & 
%  148.47
% & 
% 0.61

%     &
%     149.08\\ 
 \hline
\end{tabular}
\end{center}
\vspace{-5mm}
\label{tab:compute}
\end{table}

\subsection{Language-Conditioned Grasping}
\label{sec:expt-tog}

We now demonstrate that the single-view pose and shape estimation methods can be used for language-conditioned grasping and (qualitatively) show comparable performance with state-of-the-art baselines that use multi-view input. 
\lerfToGo~\cite{Rashid23corl-lerfToGo} is a recent method that demonstrates language-conditioned grasp synthesis for tabletop scenes. It first constructs a LERF using multi-view images and then localizes grasp poses using text-queries. 
In our modular approach for this experiment, we use the pose and shape estimation method as the first module to reconstruct the scene, and the \lerfToGo as the second module to sample language-conditioned grasps, instead of the antipodal grasp sampling. 
The object pose and shape estimation method first reconstructs a digital twin of the tabletop scene. We use \sceneCompleteShort and \samThreeD for this experiment.
We then render multiple views of the digital twin from a pre-defined camera poses and feed it into our second module, \ie the \lerfToGo pipeline. 
Given only a single-view RGB-D image, this modular method is able to generate grasps on the scene, conditioned on language quaries about the object and its parts. 
% This modular method can generate grasps on the scene, conditioned on language queries about the object and its part. 

Figure~\ref{fig:lerf} shows grasps generated using the modular approach are comparable to the baseline, both in terms of stability and being semantically meaningful. This shows the promise for robust language-conditioned grasp synthesis from single-view RGB-D images. 
This shows that the reconstructions obtained from \sceneCompleteShort are not just geometrically plausible, ensuring stable grasps, but the textures generated by diffusion-based single-view reconstruction models like InstantMesh~\cite{Xu24arxiv-instantmesh} can be used for embedding semantically meaningful features from CLIP or DINO, onto a NeRF~\cite{Rashid23corl-lerfToGo}.

\section{Conclusion}
\label{sec:conclusion}

Our work investigated the question whether recent progress in object pose and shape estimation can be leveraged for the task of grasp synthesis. In the year 2021, the answer to this question was a strong negative as Sundermeyer \etal~\cite{Sundermeyer21icra-contactGraspNet} wrote:
\begin{quote}
	``A complete 3D reconstruction enables traditional grasp planning. However, learned single-view reconstructions are often ambiguous, coarse and require class conditioning. Multiple views for 3D scanning are beneficial but not always obtainable, take additional time and typically assume a static scene.''
\end{quote}
Using \samThreeD, \crisp, and \sceneCompleteShort as the object / scene reconstruction methods and \anyGrasp as the end-to-end grasp synthesis method, we showed that the modular methods --- which first estimates object pose and shape and then samples grasps --- outperform the end-to-end method in collision-free grasp generation, force closure, and grasp stability.
Their performance remains sensitive to pose and shape estimation errors and occlusions in cluttered scenes, and is limited by run-time requirements of both, the pose and shape estimation and antipodal sampling. We demonstrated that the single-view pose and shape estimation methods can be integrated with vision-language models for language-conditioned grasping, achieving qualitatively comparable results to a multi‑view baseline.

Our results indicate that we are at an inflection point, and that the advances in object pose and shape estimation have began to show advantages to the downstream tasks like grasping. 

\bibliographystyle{ieeetr}
% \bibliography{example_paper}
%\bibliographystyle{apalike}
%\bibliographystyle{plainnat}
\bibliography{bib/refs.bib} 

@string{cvpr = {IEEE Conf. on Computer Vision and Pattern Recognition (CVPR)}}

@string{eccv = {European Conf. on Computer Vision (ECCV)}}

@string{iccv = {Intl. Conf. on Computer Vision (ICCV)}}

@string{icra = {IEEE Intl. Conf. on Robotics and Automation (ICRA)}}

@string{ijrr = {Intl. J. of Robotics Research}}

@string{iros = {IEEE/RSJ Intl. Conf. on Intelligent Robots and Systems (IROS)}}

@string{isrr = {Proc. of the Intl. Symp. of Robotics Research (ISRR)}}

@string{ral = {{IEEE} Robotics and Automation Letters ({RA-L})}}

@string{neurips = {Conf. on Neural Information Processing Systems (NeurIPS)}}

@string{pami = {{IEEE} Trans. Pattern Anal. Machine Intell.}}

@string{rss = {Robotics: Science and Systems (RSS)}}

@string{siggraph = {SIGGRAPH}}

@string{springer = {Springer Verlag}}

@string{tro = {{IEEE} Trans. Robotics}}

@inproceedings{Shi25cvpr-CRISP,
  author = {J. Shi and R. Talak and H. Zhang and D. Jin and L. Carlone},
  title = {{CRISP}: Object Pose and Shape Estimation with Test-Time Adaptation},
  booktitle = cvpr,
  note = {\award{highlight (top 13.5\%)}},
  funding = {DCIST,ONRRAPID,NSFCAREER},
  year={2025}
}

@ARTICLE{Song20ral-graspingWild,
  author={Song, Shuran and Zeng, Andy and Lee, Johnny and Funkhouser, Thomas},
  journal=ral, 
  title={Grasping in the Wild: Learning {6DoF} Closed-Loop Grasping From Low-Cost Demonstrations}, 
  year={2020},
  volume={5},
  number={3},
  pages={4978-4985} 

}

@INPROCEEDINGS{Gou21icra-rgbMatters,
  author={Gou, Minghao and Fang, Hao-Shu and Zhu, Zhanda and Xu, Sheng and Wang, Chenxi and Lu, Cewu},
  booktitle=icra, 
  title={{RGB Matters}: Learning {7-DoF} Grasp Poses on Monocular RGBD Images}, 
  year={2021}
}

@INPROCEEDINGS{Fang20cvpr-graspnet1B,
  author={Fang, Hao-Shu and Wang, Chenxi and Gou, Minghao and Lu, Cewu},
  booktitle=cvpr, 
  title={{GraspNet-1Billion}: A Large-Scale Benchmark for General Object Grasping}, 
  year={2020}
}

@article{Shimoga96ijrr-robustGraspSurvey,
author = {K.B. Shimoga},
title ={Robot Grasp Synthesis Algorithms: A Survey},
journal = ijrr,
volume = {15},
number = {3},
pages = {230-266},
year = {1996}
}

@INPROCEEDINGS{Bicchi00icra-graspingReview,
  author={Bicchi, A. and Kumar, V.},
  booktitle=icra, 
  title={Robotic grasping and contact: a review}, 
  year={2000}
}

@ARTICLE{Bohg14tro-graspSurvey,
  author={Bohg, Jeannette and Morales, Antonio and Asfour, Tamim and Kragic, Danica},
  journal=tro, 
  title={Data-Driven Grasp Synthesis—A Survey}, 
  year={2014},
  volume={30},
  number={2},
  pages={289-309}
}

@inproceedings{Eppner19isrr-billionWaysToGrasp,
    title = {A Billion Ways to Grasps - An Evaluation of Grasp Sampling Schemes on a Dense, Physics-based Grasp Data Set},
    author = {Clemens Eppner and Arsalan Mousavian and Dieter Fox},
    year = {2019},
    booktitle = isrr
}

@article{Mahler17rss-dexnet2,
  title={Dex-Net 2.0: Deep Learning to Plan Robust Grasps with Synthetic Point Clouds and Analytic Grasp Metrics},
  author={Mahler, Jeffrey and Liang, Jacky and Niyaz, Sherdil and Laskey, Michael and Doan, Richard and Liu, Xinyu and Ojea, Juan Aparicio and Goldberg, Ken},
  booktitle=rss,
  year={2017}
}

@article{Zhai22ral-da2dataset,
  author={Zhai, Guangyao and Zheng, Yu and Xu, Ziwei and Kong, Xin and Liu, Yong and Busam, Benjamin and Ren, Yi and Navab, Nassir and Zhang, Zhengyou},
  journal=ral, 
  title={DA$^2$ Dataset: Toward Dexterity-Aware Dual-Arm Grasping}, 
  year={2022},
  volume={7},
  number={4},
  pages={8941-8948} 
}

@article{Sundermeyer21icra-contactGraspNet,
  title={Contact-GraspNet: Efficient 6-DoF Grasp Generation in Cluttered Scenes},
  author={Sundermeyer, Martin and Mousavian, Arsalan and Triebel, Rudolph and Fox, Dieter},
  booktitle=icra,
  year={2021}
}

@ARTICLE{Newbury23tro-graspReview,
  author={Newbury, Rhys and Gu, Morris and Chumbley, Lachlan and Mousavian, Arsalan and Eppner, Clemens and Leitner, Jürgen and Bohg, Jeannette and Morales, Antonio and Asfour, Tamim and Kragic, Danica and Fox, Dieter and Cosgun, Akansel},
  journal=tro, 
  title={Deep Learning Approaches to Grasp Synthesis: A Review}, 
  year={2023},
  volume={39},
  number={5},
  month = {Jun.},
  pages={3994-4015}
}

@ARTICLE{Fang23tro-anyGrasp,
  author={Fang, Hao-Shu and Wang, Chenxi and Fang, Hongjie and Gou, Minghao and Liu, Jirong and Yan, Hengxu and Liu, Wenhai and Xie, Yichen and Lu, Cewu},
  journal=tro, 
  title={AnyGrasp: Robust and Efficient Grasp Perception in Spatial and Temporal Domains}, 
  year={2023},
  volume={39},
  number={5},
  pages={3929-3945},
  month = {Jun.}
}

@inproceedings{Rashid23corl-lerfToGo,
 title={Language Embedded Radiance Fields for Zero-Shot Task-Oriented Grasping},
 author={Adam Rashid and Satvik Sharma and Chung Min Kim and Justin Kerr and Lawrence Yunliang Chen and Angjoo Kanazawa and Ken Goldberg},
 booktitle=corl,
 year = {2023}
}

@article{Fang20ijrr-tog,
author = {Kuan Fang and Yuke Zhu and Animesh Garg and Andrey Kurenkov and Viraj Mehta and Li Fei-Fei and Silvio Savarese},
title ={Learning task-oriented grasping for tool manipulation from simulated self-supervision},
journal = ijrr,
volume = {39},
number = {2-3},
pages = {202-216},
year = {2020}
}

@INPROCEEDINGS{Vuong24cvpr-languageGrasp,
  author={Vuong, An Dinh and Vu, Minh Nhat and Huang, Baoru and Nguyen, Nghia and Le, Hieu and Vo, Thieu and Nguyen, Anh},
  booktitle=cvpr, 
  title={Language-driven Grasp Detection}, 
  year={2024}  
  }

@inproceedings{Wei24icra-droGrasp,
    title={{D(R,O) Grasp}: A Unified Representation of Robot and Object Interaction for Cross-Embodiment Dexterous Grasping},
    author={Wei, Zhenyu and Xu, Zhixuan and Guo, Jingxiang and Hou, Yiwen and Gao, Chongkai and Cai, Zhehao and Luo, Jiayu and Shao, Lin},
    booktitle=icra,
    year={2025} 
}

@article{Agarwal24arxiv-sceneComplete,
  title = {{{SceneComplete}}: {{Open-World 3D Scene Completion}} in {{Complex Real World Environments}} for {{Robot Manipulation}}}, 
  author = {Agarwal, Aditya and Singh, Gaurav and Sen, Bipasha and Lozano-Pérez, Tomás and Kaelbling, Leslie Pack},
  year = {2024},
  month = {Oct.}, 
  journal = {arXiv preprint arxiv:2410.23643},
  url = {http://arxiv.org/abs/2410.23643}
}

@article{Watson22arxiv-novelView,
      title={Novel View Synthesis with Diffusion Models}, 
      author={Daniel Watson and William Chan and Ricardo Martin-Brualla and Jonathan Ho and Andrea Tagliasacchi and Mohammad Norouzi},
      year={2022},
      journal = {arxiv preprints arxiv:2210.04628}
}

@article{Fu22nips-wild6d,
  title={Category-level 6d object pose estimation in the wild: A semi-supervised learning approach and a new dataset},
  author={Fu, Yang and Wang, Xiaolong},
  journal=neurips,
  volume={35},
  pages={27469--27483},
  year={2022}
}

@inproceedings{Chen20iccv-learningCanonicalShape,
  title={Learning canonical shape space for category-level 6d object pose and size estimation},
  author={Chen, Dengsheng and Li, Jun and Wang, Zheng and Xu, Kai},
  booktitle=cvpr,
  pages={11973--11982},
  year={2020}
}

@inproceedings{Chen21iccv-sgpa,
	title={Sgpa: Structure-guided prior adaptation for category-level 6d object pose estimation},
	author={Chen, Kai and Dou, Qi},
	booktitle=iccv,
	pages={2773--2782},
	year={2021}
}

@inproceedings{Tian20eccv-SPD,
	title={Shape prior deformation for categorical 6d object pose and size estimation},
	author={Tian, Meng and Ang, Marcelo H and Lee, Gim Hee},
	booktitle=eccv,
	pages={530--546},
	year={2020},
	organization={Springer}
}

@inproceedings{Lin21iccv-dualposenet,
  title={Dualposenet: Category-level 6d object pose and size estimation using dual pose network with refined learning of pose consistency},
  author={Lin, Jiehong and Wei, Zewei and Li, Zhihao and Xu, Songcen and Jia, Kui and Li, Yuanqing},
  booktitle=iccv,
  pages={3560--3569},
  year={2021}
}

@inproceedings{Lunayach24icra-FSD,
  title={Fsd: Fast self-supervised single rgb-d to categorical 3d objects},
  author={Lunayach, Mayank and Zakharov, Sergey and Chen, Dian and Ambrus, Rares and Kira, Zsolt and Irshad, Muhammad Zubair},
  booktitle=icra,
  pages={14630--14637},
  year={2024},
  organization={IEEE}
}

@inproceedings{Hong24iclr-lrm,
title={{LRM}: Large Reconstruction Model for Single Image to 3D},
author={Yicong Hong and Kai Zhang and Jiuxiang Gu and Sai Bi and Yang Zhou and Difan Liu and Feng Liu and Kalyan Sunkavalli and Trung Bui and Hao Tan},
booktitle=iclr,
year={2024},
url={https://openreview.net/forum?id=sllU8vvsFF}
}

@inproceedings{Huang24cvpr-zeroShape,
  author    = {Huang, Zixuan and Stojanov, Stefan and Thai, Anh and Jampani, Varun and Rehg, James M},
  title     = {ZeroShape: Regression-based Zero-shot Shape Reconstruction},
  booktitle = cvpr, 
  year      = {2024},
}

@INPROCEEDINGS{Wu23cvpr-mvCompressiveCoding,
  author={Wu, Chao-Yuan and Johnson, Justin and Malik, Jitendra and Feichtenhofer, Christoph and Gkioxari, Georgia},
  booktitle=cvpr, 
  title={Multiview Compressive Coding for 3D Reconstruction}, 
  year={2023} 
}

@inproceedings{Alwala22cvpr-SS3D,
  title={Pre-train, Self-train, Distill: A simple recipe for Supersizing 3D Reconstruction},
  author={Vasudev, Kalyan Alwala and  Gupta, Abhinav and Tulsiani, Shubham},
  year={2022},
  booktitle=cvpr
}

@article{Xu24arxiv-instantmesh,
  title={InstantMesh: Efficient 3D Mesh Generation from a Single Image with Sparse-view Large Reconstruction Models},
  author={Xu, Jiale and Cheng, Weihao and Gao, Yiming and Wang, Xintao and Gao, Shenghua and Shan, Ying},
  journal={arXiv preprint arXiv:2404.07191},
  year={2024}
}

@INPROCEEDINGS{Liu23iccv-zero1to3,
  author={Liu, Ruoshi and Wu, Rundi and Van Hoorick, Basile and Tokmakov, Pavel and Zakharov, Sergey and Vondrick, Carl},
  booktitle=iccv, 
  title={Zero-1-to-3: Zero-shot One Image to 3D Object}, 
  year={2023}
}

@article{Jun23arxiv-shape,
  title={Shap-e: Generating conditional 3d implicit functions},
  author={Jun, Heewoo and Nichol, Alex},
  journal={arXiv preprint arXiv:2305.02463},
  year={2023}
}

@INPROCEEDINGS{Xu23cvpr-neuralLift360,
  author={Xu, Dejia and Jiang, Yifan and Wang, Peihao and Fan, Zhiwen and Wang, Yi and Wang, Zhangyang},
  booktitle=cvpr, 
  title={{NeuralLift-360}: Lifting an in-the-Wild 2D Photo to A 3D Object with 360° Views}, 
  year={2023}
}

@inproceedings{MelasKyriazi23cvpr-realFusion,
  author = {Melas-Kyriazi, Luke and Rupprecht, Christian and Laina, Iro and Vedaldi, Andrea},
  title = {{RealFusion}: 360 Reconstruction of Any Object from a Single Image},
  booktitle=cvpr,
  year = {2023},
  url = {https://arxiv.org/abs/2302.10663},
}

@inproceedings{Wang21cvpr-GDRNetGeometryGuided,
	title = {{{GDR-Net}}: {{Geometry-Guided Direct Regression Network}} for {{Monocular 6D Object Pose Estimation}}}, 
	booktitle = cvpr, 
	author = {Wang, Gu and Manhardt, Fabian and Tombari, Federico and Ji, Xiangyang},
	year = {2021},
	pages = {16611--16621}
}

@inproceedings{Wang19cvpr-nocs,
	title={Normalized object coordinate space for category-level 6d object pose and size estimation},
	author={H. Wang and S. Sridhar and J. Huang and J. Valentin and S. Song and L. Guibas},
	booktitle=cvpr,
	pages={2642--2651},
	year={2019}
}

@ARTICLE{Liu25pami-GDRNPP,
  author={Liu, Xingyu and Zhang, Ruida and Zhang, Chenyangguang and Wang, Gu and Tang, Jiwen and Li, Zhigang and Ji, Xiangyang},
  journal=pami, 
  title={{GDRNPP}: A Geometry-Guided and Fully Learning-Based Object Pose Estimator}, 
  year={2025},
  volume={47},
  number={7},
  pages={5742-5759}
}

@INPROCEEDINGS{Sen23icra-scarpShapeCompletion,
  author={Sen, Bipasha and Agarwal, Aditya and Singh, Gaurav and B., Brojeshwar and Sridhar, Srinath and Krishna, Madhava},
  booktitle=icra, 
  title={SCARP: 3D Shape Completion in ARbitrary Poses for Improved Grasping}, 
  month = {Jun.},
  year={2023}
}

@INPROCEEDINGS{Wu23rss-graspSuperquadrics, 
    AUTHOR    = {Yuwei Wu AND Weixiao Liu AND Zhiyang Liu AND Gregory S Chirikjian}, 
    TITLE     = {{Learning-Free Grasping of Unknown Objects Using Hidden Superquadrics}}, 
    BOOKTITLE = rss, 
    YEAR      = {2023},  
    MONTH     = {Jul.}
}

@ARTICLE{Chisari24ral-centerGrasp,
  author={Chisari, Eugenio and Heppert, Nick and Welschehold, Tim and Burgard, Wolfram and Valada, Abhinav},
  journal=ral, 
  title={CenterGrasp: Object-Aware Implicit Representation Learning for Simultaneous Shape Reconstruction and 6-DoF Grasp Estimation}, 
  year={2024},
  volume={9},
  number={6},
  pages={5094-5101}
}

@inproceedings{Xiang18rss-posecnn,
	title={{PoseCNN}: A convolutional neural network for {6D} object pose estimation in cluttered scenes},
	author={Xiang, Yu and Schmidt, Tanner and Narayanan, Venkatraman and Fox, Dieter},
	booktitle=rss,
	year={2018}
}

@InProceedings{Wang20eccv-Self6DSelfSupervised,
	author="Wang, Gu
	and Manhardt, Fabian
	and Shao, Jianzhun
	and Ji, Xiangyang
	and Navab, Nassir
	and Tombari, Federico",
	editor="Vedaldi, Andrea
	and Bischof, Horst
	and Brox, Thomas
	and Frahm, Jan-Michael",
	title="{Self6D}: Self-supervised Monocular {6D} Object Pose Estimation",
	booktitle=eccv,
	year="2020",
	month={Nov.},
	pages="108--125"
}

@misc{Zhang24arxiv-shapeICP,
      title={{ShapeICP}: Iterative Category-level Object Pose and Shape Estimation from Depth}, 
      author={Y. Zhang and J.J. Leonard},
      year={2024},
      eprint={2408.13147},
      archivePrefix={arXiv},
      primaryClass={cs.CV},
      myNotes = {dense active shape model on meshes -- very interesting idea to deform each object from a default template. 
      Then ICP-style alternation of pose and shape estimation, with the shape estimation being done via gradient descent on a complex loss. 
      Results are mixed and they propose many heuristics to improve convergence which suggests that the problem is very hard.},
      url={https://arxiv.org/abs/2408.13147}, 
}

@inproceedings{Caraffa24eccv-freeze, 
  title={{FreeZe}: Training-free zero-shot 6D pose estimation with geometric and vision foundation models}, 
  author = {Caraffa, Andrea and Boscaini, Davide and Hamza, Amir and Poiesi, Fabio}, 
  booktitle = eccv, 
  year = {2024} 
}

@inproceedings{Moon25cvpr-coop,
    title={{Co-op}: Correspondence-based Novel Object Pose Estimation},
    author={Moon, Sungphill and Son, Hyeontae and Hur, Dongcheol and Kim, Sangwook},
    booktitle=cvpr,
    year={2025}
}

@InProceedings{Wen24cvpr-foundationPose,
author        = {Bowen Wen, Wei Yang, Jan Kautz and Stan Birchfield},
title         = {{FoundationPose}: Unified 6D Pose Estimation and Tracking of Novel Objects},
booktitle     = cvpr,
year          = {2024},
}

@inproceedings{Nguyen24cvpr-gigaPose,
    title={{GigaPose}: Fast and Robust Novel Object Pose Estimation via One Correspondence},
    author={Nguyen, Van Nguyen and Groueix, Thibault and Salzmann, Mathieu and Lepetit, Vincent},
    booktitle = cvpr,
    year={2024}
}

@InProceedings{Moon24cvpr-genFlow,
    author    = {Moon, Sungphill and Son, Hyeontae and Hur, Dongcheol and Kim, Sangwook},
    title     = {{GenFlow}: Generalizable Recurrent Flow for 6D Pose Refinement of Novel Objects},
    booktitle = cvpr,
    year      = {2024} 
}

@inproceedings{Labbe22corl-megapose,
  title={Megapose: 6d pose estimation of novel objects via render \& compare},
  author={Labb{\'e}, Yann and Manuelli, Lucas and Mousavian, Arsalan and Tyree, Stephen and Birchfield, Stan and Tremblay, Jonathan and Carpentier, Justin and Aubry, Mathieu and Fox, Dieter and Sivic, Josef},
  journal=corl,
  year={2022}
}

@article{Hodan20eccvw-BOPChallenge,
title={{BOP} Challenge 2020 on {6D} Object Localization},
author={Hoda{\v{n}}, Tom{\'a}{\v{s}} and Sundermeyer, Martin and Drost, Bertram and Labb{\'e}, Yann and Brachmann, Eric and Michel, Frank and Rother, Carsten and Matas, Ji{\v{r}}{\'i}},
journal={European Conference on Computer Vision Workshops (ECCVW)},
year={2020}
}

@ARTICLE{Miller04ram-graspIt,
  author={Miller, A.T. and Allen, P.K.},
  journal=ram, 
  title={{Graspit!} A versatile simulator for robotic grasping}, 
  year={2004},
  volume={11},
  number={4},
  pages={110-122}
}

@inproceedings{Lorensen87siggraph-marchingCubes,
	Author = {W.E. Lorensen and H.E. Cline},
	Booktitle = SIGGRAPH,
	pages = {163--169},
	Title = {Marching cubes: A high resolution 3d surface construction algorithm},
	Year = 1987
	}

@article{Ju24arxiv-brushNet,
  title={{BrushNet}: A Plug-and-Play Image Inpainting Model with Decomposed Dual-Branch Diffusion}, 
  author={Xuan Ju and Xian Liu and Xintao Wang and Yuxuan Bian and Ying Shan and Qiang Xu},
  year={2024},
  journal={arxiv pre-print arxiv:2403.06976} 
}

@INPROCEEDINGS{Gao21icra-shapeCompletionGrasp,
  author={Gao, Wei and Tedrake, Russ},
  booktitle=icra, 
  title={{kPAM-SC}: Generalizable Manipulation Planning using KeyPoint Affordance and Shape Completion}, 
  year={2021}, 
  pages={6527-6533} 
}

@ARTICLE{Kiatos20tro-shapeCompletionGrasp,
  author={Kiatos, Marios and Malassiotis, Sotiris and Sarantopoulos, Iason},
  journal=tro, 
  title={A Geometric Approach for Grasping Unknown Objects With Multifingered Hands}, 
  year={2021},
  volume={37},
  number={3},
  pages={735-746}
}

@INPROCEEDINGS{ChavanDafle22iros-shapeCompletionGrasp,
  author={Chavan-Dafle, Nikhil and Popovych, Sergiy and Agrawal, Shubham and Lee, Daniel D. and Isler, Volkan},
  booktitle=iros, 
  title={Simultaneous Object Reconstruction and Grasp Prediction using a Camera-centric Object Shell Representation}, 
  year={2022}, 
  pages={1396-1403} 
}

@ARTICLE{Gualtieri21ral-shapeCompletionGrasp,
  author={Gualtieri, Marcus and Platt, Robert},
  journal=ral, 
  title={Robotic Pick-and-Place With Uncertain Object Instance Segmentation and Shape Completion}, 
  year={2021},
  volume={6},
  number={2},
  pages={1753-1760}
}

@INPROCEEDINGS{Lundell19iros-shapeCompletionGrasp,
  author={Lundell, Jens and Verdoja, Francesco and Kyrki, Ville},
  booktitle=iros, 
  title={Robust Grasp Planning Over Uncertain Shape Completions}, 
  year={2019}, 
  pages={1526-1532}
}

@INPROCEEDINGS{Jiang21rss-shapeCompletionAuxGrasp, 
    AUTHOR    = {Zhenyu Jiang AND Yifeng Zhu AND Maxwell Svetlik AND Kuan Fang AND Yuke Zhu}, 
    TITLE     = {{Synergies Between Affordance and Geometry: 6-DoF Grasp Detection via Implicit Representations}}, 
    BOOKTITLE = rss, 
    YEAR      = {2021}
}

@INPROCEEDINGS{Yan18icra-shapeCompletionAuxGrasp,
  author={Yan, Xinchen and Hsu, Jasmined and Khansari, Mohammad and Bai, Yunfei and Pathak, Arkanath and Gupta, Abhinav and Davidson, James and Lee, Honglak},
  booktitle=icra, 
  title={Learning 6-DOF Grasping Interaction via Deep Geometry-Aware 3D Representations}, 
  year={2018}
}

@INPROCEEDINGS{Yang21icra-shapeCompletionAuxGrasp,
  author={Yang, Daniel and Tosun, Tarik and Eisner, Benjamin and Isler, Volkan and Lee, Daniel},
  booktitle=icra, 
  title={Robotic Grasping through Combined Image-Based Grasp Proposal and 3D Reconstruction}, 
  year={2021}
}

@Misc{pybullet24,
  title = {{PyBullet}: a Python module for physics simulation for games, robotics and machine learning},
  howpublished = {\url{http://pybullet.org}},
  year = {2024}  
}

@article{sam3d-objects,
      title={SAM 3D: 3Dfy Anything in Images}, 
      author={SAM 3D Team and Xingyu Chen and Fu-Jen Chu and Pierre Gleize and Kevin J Liang and Alexander Sax and Hao Tang and Weiyao Wang and Michelle Guo and Thibaut Hardin and Xiang Li and Aohan Lin and Jiawei Liu and Ziqi Ma and Anushka Sagar and Bowen Song and Xiaodong Wang and Jianing Yang and Bowen Zhang and Piotr Dollár and Georgia Gkioxari and Matt Feiszli and Jitendra Malik},
      year={2025},
      journal={arxiv pre-print arxiv:2511.16624} 
}

% \appendix
% \input{sections/app-modular-methods.tex}
% \input{sections/app-eval-metrics.tex}
% \input{sections/app-expt-setup.tex}
% \input{sections/app-failure-analysis.tex}

\end{document}